\documentclass{article}

\usepackage[final,nonatbib]{neurips_2024}

\usepackage[utf8]{inputenc} %
\usepackage[T1]{fontenc}    %
\usepackage{hyperref}       %
\usepackage{url}            %
\usepackage{booktabs}       %
\usepackage{amsfonts}       %
\usepackage{nicefrac}       %
\usepackage{microtype}      %
\usepackage{xcolor}         %
\usepackage{amsmath}
\usepackage{bm}
\usepackage{multirow}
\usepackage{graphicx}
\usepackage{xspace}
\usepackage{wrapfig}
\usepackage{colortbl}
\usepackage{todonotes}
\usepackage{mathrsfs}
\usepackage{makecell}
\usepackage{lipsum}
\usepackage{enumitem}

\usepackage[capitalize]{cleveref}
\crefname{section}{Sec.}{Secs.}
\Crefname{section}{Section}{Sections}
\Crefname{table}{Table}{Tables}
\crefname{table}{Tab.}{Tabs.}

\makeatletter
\DeclareRobustCommand\onedot{\futurelet\@let@token\@onedot}
\def\@onedot{\ifx\@let@token.\else.\null\fi\xspace}

\def\eg{\emph{e.g}\onedot} 
\def\ie{\emph{i.e}\onedot}

\makeatother

\definecolor{mygray}{gray}{0.9}  %
\definecolor{ode_green}{RGB}{28,175,88}  %
\definecolor{grad1}{HTML}{d2eaeb}
\definecolor{grad2}{HTML}{98d7d0}
\definecolor{grad3}{HTML}{4dabb1}
\definecolor{cvprpink}{RGB}{219, 48, 122}

\newcommand{\MethodFull}{Motion Consistency Model\xspace}  %
\newcommand{\methodfull}{motion consistency model\xspace}  %
\newcommand{\methodabbr}{MCM\xspace}

\hypersetup{
    colorlinks=true,
    linkcolor=cvprpink,
    urlcolor=cvprpink,
}

\title{\MethodFull:\\Accelerating Video Diffusion with \\Disentangled
Motion-Appearance Distillation}

\author{%
Yuanhao Zhai$^{*\dagger}$~~~
Kevin Lin$^{\ddagger}$~~~
Zhengyuan Yang$^{\ddagger}$~~~
Linjie Li$^{\ddagger}$~~~
Jianfeng Wang$^{\ddagger}$~~~\\
\textbf{Chung-Ching Lin}$^{\ddagger}$~~~
\textbf{David Doermann}$^{\dagger}$~~~
\textbf{Junsong Yuan}$^{\dagger}$~~~
\textbf{Lijuan Wang}$^{\ddagger}$\\
\normalsize$^\dagger$State University of New York at Buffalo~~~
$^\ddagger$Microsoft\\
\footnotesize{\texttt{\{yzhai6,doermann,jsyuan\}@buffalo.edu}}~~~\\
\footnotesize{\texttt{\{keli,zhengyang,lindsey.li,jianfw,chungching.lin,lijuanw\}@microsoft.com}}~~~
\\[0.1cm]
{\url{https://yhzhai.github.io/mcm/}}
\vspace{-.1in}
}

\begin{document}

\maketitle

\begin{abstract}
Image diffusion distillation achieves high-fidelity generation with very few sampling steps. However, applying these techniques directly to video diffusion often results in unsatisfactory frame quality 
due to the limited visual quality in public video datasets. This affects the performance of both teacher and student video diffusion models.
Our study aims to improve video diffusion distillation while improving frame appearance using abundant high-quality image data. 
We propose \methodfull (\methodabbr), a single-stage video diffusion distillation method that disentangles motion and appearance learning. Specifically, \methodabbr includes a video consistency model that distills motion from the video teacher model, and an image discriminator that enhances frame appearance to match high-quality image data. 
This combination presents two challenges:
(1) conflicting frame learning objectives, as video distillation learns from low-quality video frames while the image discriminator targets high-quality images; and (2) training-inference discrepancies due to the differing quality of video samples used during training and inference. 
To address these challenges, we introduce disentangled motion distillation and
mixed trajectory distillation.
The former applies the distillation objective solely to the motion
representation, while the latter mitigates training-inference discrepancies by
mixing distillation trajectories from both the low- and high-quality %
video domains.
Extensive experiments show that our \methodabbr achieves the state-of-the-art
video diffusion distillation performance.
Additionally, our method can enhance frame quality in video diffusion models,
producing frames with high aesthetic scores or specific styles without corresponding video data.

\end{abstract}

\begingroup
\renewcommand{\thefootnote}{}
\footnotetext{$^*$Work done during an internship at Microsoft.}
\addtocounter{footnote}{-1}
\endgroup

\begin{figure}[t]
    \centering
    \includegraphics[width=\textwidth]{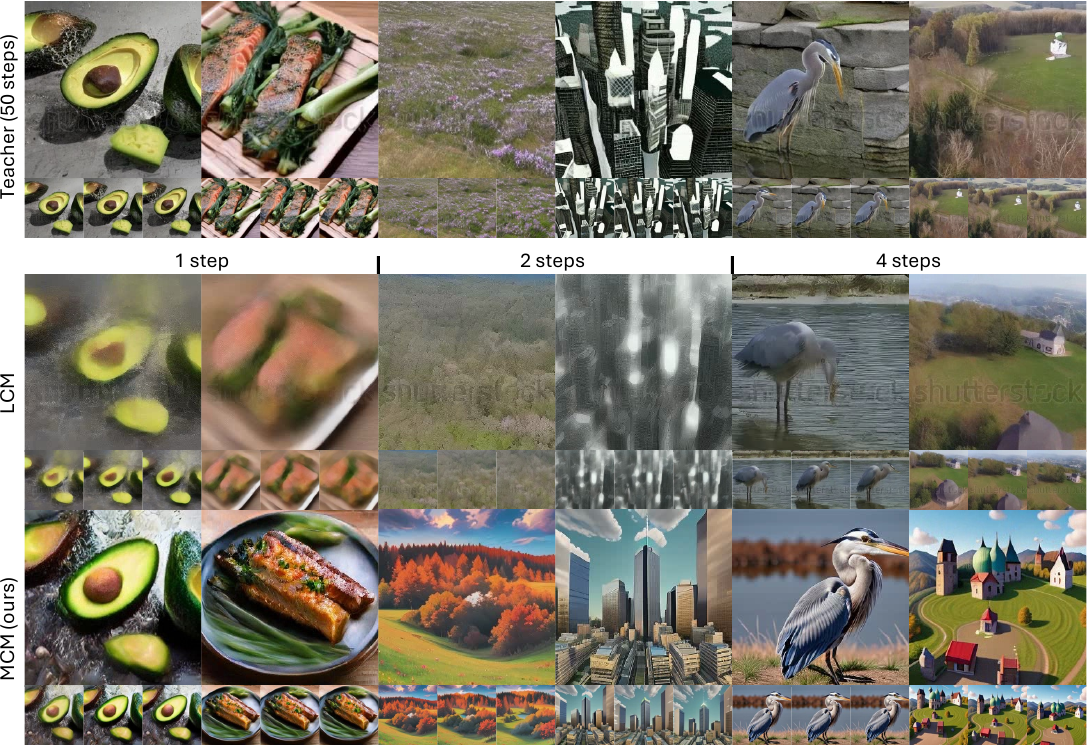}
    \caption{Qualitative result comparisons with latent consistency model
    (LCM)~\cite{luo2023latent} using ModelScopeT2V~\cite{wang2023modelscope} as
    teacher. Our \methodabbr outputs clearer video frames using few-step
    sampling, and can also adapt to different image styles using additional
    image datasets. Corresponding video results are in supplementary materials.}
    \label{fig:teaser}
\end{figure}

\section{Introduction}

Diffusion
models~\cite{ho2020denoising,song2020denoising,nichol2021improved,ho2022classifier,karras2022elucidating,rombach2022high}
have significantly advanced the quality of text-to-image
generation~\cite{rombach2022high,betker2023improving,dai2023emu}, enabling the
creation of high-fidelity images and allowing for diverse stylistic
variations~\cite{toonyou2023,realisticvisionv60b1,disneypixarcartoonb}.
The recent development in image diffusion distillation~\cite{song2023consistency,lin2024sdxl,sauer2023adversarial,luo2023latent,kohler2024imagine} significantly reduces the inference cost by letting distilled student models to perform high-fidelity generation with extremely low sampling steps. 

Due to the additional temporal dimension, the sampling time for video
diffusion models~\cite{ho2022imagen,ho2022video,blattmann2023align,wang2023modelscope,guo2023animatediff,wang2024videocomposer} is considerably longer than that for image diffusion models.
However, directly extending image distillation methods to speed up the video diffusion models results in unsatisfactory frame appearance quality.
Specifically, unlike image diffusion models that benefit from high-quality image
datasets such as LAION~\cite{schuhmann2022laion}, public video
datasets~\cite{xu2016msr,soomro2012ucf101,bain2021frozen} often suffer from lower frame quality, 
including issues like low resolution, significant motion blur, and watermarks.
This discrepancy poses a great challenge, as the quality of the training data
directly impacts both the teacher generation quality and the distillation process.
An example is shown in~\cref{fig:teaser}, where both the teacher and the latent 
consistency model (LCM) student~\cite{luo2023latent,wang2023videolcm} learn the watermark from the video dataset.

\begin{figure}[t]
    \centering
    \includegraphics[width=\textwidth]{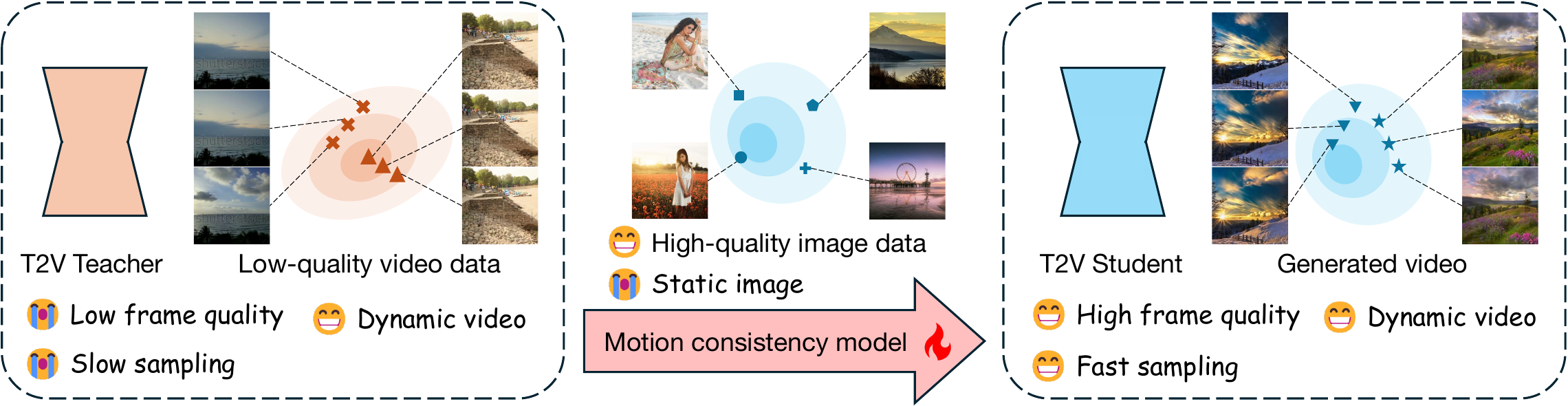}
    \caption{Illustration of our \methodfull distillation process, which not only distills the motion
    prior from teacher to accelerate sampling, but also can benefit from an
    additional high-quality image dataset to improve the frame quality of
    generated videos.}
    \label{fig:task-illustration}
\end{figure}

To address this challenge, we propose \methodfull (\methodabbr), a single-stage
video diffusion distillation method that enables few-step sampling and
can optionally leverage a high-quality \emph{image} dataset to improve the
video frame quality.
We showcase our results in~\cref{fig:teaser} bottom row and \cref{fig:appearance-adjustment}, and illustrate the distillation process in~\cref{fig:task-illustration}.

To simultaneously achieve the goal of diffusion distillation and frame quality
improvement, we build our method upon a video LCM~\cite{luo2023latent,wang2023videolcm} combined with an image adversarial
objective~\cite{sauer2023stylegan,sauer2023adversarial}, due to their proven
effectiveness in the two
tasks~\cite{luo2023latent,wang2023videolcm,wang2024animatelcm,zhu2017unpaired,isola2017image},
respectively.
However, directly combining them presents two essential problems.
(1)~\emph{Conflicted frame learning objectives}.
The LCM aims to learn %
and represent the characteristics of low-quality video frames, while the adversarial objective pushes the frames toward high-quality image data.
This opposition creates conflicts between preserving low-quality visual features and enhancing image quality.
(2)~\emph{Discrepant training-inference distillation input}.
During training, only low-quality video samples are used for LCM distillation,
whereas during inference, the input is generated high-quality video samples.
These two problems greatly hinders the distillation process and the generated
frame quality.

To address the first problem, we propose disentangled motion distillation.
Specifically, we disentangle the motion components from the video latent, and
only distill the motion knowledge from the teacher.
Meanwhile, the image adversarial objective is used exclusively to learn frame
appearance.
This approach effectively mitigates the conflict between learning objectives by
ensuring that motion and appearance are learned independently.
For the second problem, we propose mixed trajectory distillation.
We simulate the probability flow ordinary differential equation (PF-ODE)
trajectory during inference and sample generated high-quality videos from this
simulated trajectory for distillation.
By mixing distillation between low-quality video samples and generated
high-quality video samples, we ensure better alignment between training and
inference phases, thereby enhancing overall performance.

In summary, our contributions are as follows:
\begin{itemize}[leftmargin=*,noitemsep,topsep=0pt]
\item We propose \methodfull (MCM), a single-stage video
diffusion distillation method that accelerates the sampling process and
can optionally leverage an image dataset to enhance the quality of generated video
frames.
\item We introduce disentangled motion distillation to resolve the conflicting
frame learning objectives between video diffusion distillation and image
adversarial learning, ensuring better motion consistency and frame appearance.
\item We propose mixed trajectory distillation, which simulates the PF-ODE
trajectory during inference.
By combining low-quality video samples with generated high-quality video samples
during distillation, our approach improves training-inference alignment and enhances generation quality.
\item We conduct extensive experiments, demonstrating that our \methodabbr
significantly improves video diffusion distillation performance.
Furthermore, when leveraging an additional image dataset, our \methodabbr better
aligns the appearance of the generated video with the high-quality image
dataset.
\end{itemize}

\section{Preliminaries}

\label{sec:preliminaries}

\noindent\textbf{Diffusion models.}
Diffusion models~\cite{ho2020denoising} consists of a forward and a reverse
process.
The forward process corrupts the data progressively by interpolating a Gaussian
noise $\bm{\epsilon} \in \mathcal{N}(0, \bm{I})$ and the data sample $\bm{x}_0$:
$\bm{x}_t = q(\bm{x}_0, \bm{\epsilon}, t) = \alpha_t \bm{x}_0 + \sigma_t
\bm{\epsilon}$, where $\alpha_t$ and $\sigma_t$ are pre-defined noise schedule.
The reverse process learns a time-conditioned model to gradually removes the
noise.
Recent methods~\cite{song2020score,lu2022dpm} propose to design numerical
solvers of ordinary differential equations (ODE) to reduce the number of
sampling steps.

\noindent\textbf{Consistency model.}
Consistency model (CM)~\cite{song2023consistency} is a new family of generative
models that enables few-step sampling.
The essence of CM $\bm{f}(\bm{x}_t, t)$ is the self-consistency property, that
maps any point in the same probability flow ODE (PF-ODE) trajectory to a same
point.
That is, $\bm{f}(\bm{x}_t, t) = \bm{f}(\bm{x}_{t'}, t')$ for $\forall t,t'\in
[\epsilon, T]$, where $\epsilon$ is a fixed small positive number. 
The boundary condition $\bm{f}(\bm{x}_\epsilon, \epsilon) = \bm{x}_\epsilon$
ensures the CM always predicts the origin of PF-ODE trajectories.
CM is parameterized as the weighted sum of the data sample and a model output
$\bm{F}$ parameterized by $\theta$:
\begin{equation}
\bm{f}(\bm{x}_t, t) = c_{\text{skip}}(t) \bm{x}_t + c_{\text{out}}(t)
\bm{F}_\theta(\bm{x}_t, t),
\end{equation}
where $c_{\text{skip}}(t)$ and $c_{\text{out}}(t)$ are differentiable
functions~\cite{song2023consistency,karras2022elucidating} with
$c_{\text{skip}}(\epsilon)=1$ and $c_{\text{out}}(\epsilon)=0$.

CM can be learned via consistency distillation (CD), where a teacher
model $\phi$ and an ODE solver $\bm{\Phi}$ estimates the previous sample in the
empirical PF-ODE trajectory, with $s$ being the step size:
\begin{equation}
\hat{\bm{x}}_{t-s}^{\phi} = \bm{x}_{t} - s \bm{\Phi}(\bm{x}_{t}, t; \phi).
\end{equation}
Let $\theta$ be the student model parameter, and its exponential moving average
(EMA) $\theta^- \leftarrow \mu \theta^- + (1-\mu) \theta$ be the target model
parameter.
CD is trained to enforce the self-consistency property, which minimizes the
distance between student and target outputs:
\begin{equation}
\mathcal{L}_{\text{CD}} = \mathbb{E}_{\bm{x}_0,\bm{\epsilon},t} \left[ d \left(
\bm{f}_{\theta} (\bm{x}_{t}, t), \bm{f}_{\theta^-} (\hat{\bm{x}}_{t-s}^{\phi},
t-s) \right) \right].
\label{eq:cd}
\end{equation}
In this paper, we follow previous
methods~\cite{luo2023latent,wang2024animatelcm} to use Huber loss as the
distance function: $d(\bm{x}_t, \bm{x}_t') = \sqrt{\| \bm{x}_t - \bm{x}_t'
\|^2_2 + \delta^2} - \delta$, where $\delta$ is a threshold hyperparameter.

\section{Motion consistency model}

We present \methodfull (\methodabbr), a novel video diffusion distillation
method designed for accelerating the text-to-video diffusion sampling process.
Additionally, \methodabbr can leverage a different image dataset to adjust the
frame appearance, \eg, frame quality improvement, watermark removal or frame
style transfer.
An illustration of the objective is shown in~\cref{fig:task-illustration}.

\noindent\textbf{Problem formulation.}
Given a pre-trained text-to-video diffusion model $\phi$, a video-caption
dataset $\mathbb{V} = \{(\bm{v}, c)_i\}$, and an image dataset $\mathbb{M} = \{
m_i \}$, our objective is twofold.
\begin{enumerate}[leftmargin=*,noitemsep,topsep=0pt]
\item Distill the \emph{motion} knowledge from teacher $\phi$ into a student
$\theta$ using the video dataset $\mathbb{V}$, enabling few-step video
generation.
\item Adjust the \emph{appearance} of the generated video frames using the image
dataset $\mathbb{M}$.
\end{enumerate}
In this paper, we assume that the appearance of video frames in $\mathbb{V}$
has lower quality compared to the images in $\mathbb{M}$.
For convenience, we refer to the appearance of video frames as following a
\emph{``low-quality''} distribution and the images in $\mathbb{M}$ as following a
\emph{``high-quality''} distribution.

When the image dataset $\mathbb{M}$ consists of frames extracted from the video dataset $\mathbb{V}$, this task simplifies into a straightforward video diffusion distillation process.

\subsection{Video latent consistency model meets image discriminator}

A straightforward way to achieve the aforementioned goal is to adopt a two-stage
training strategy~\cite{wang2024animatelcm}, \ie, train a 2D backbone with the
image dataset, and then temporally inflate the backbone to fine-tune temporal
layers on videos.
However, not only this approach can be time-consuming, but the disjointed
training stages can lead to suboptimal motion modeling and degraded appearance
learning.
To address this problem, we propose a single-stage, end-to-end learnable approach.

To build a strong baseline, we adopt a video latent consistency model
(LCM)~\cite{luo2023latent} $\bm{f}_{\theta}(\cdot, \cdot)$ with a frame
adversarial objective~\cite{sauer2023stylegan,sauer2023adversarial}.
The LCM has proven effective in diffusion
distillation~\cite{luo2023latent,wang2023videolcm,wang2024animatelcm}, and the
adversarial objective has been widely used for style
transfer~\cite{zhu2017unpaired,isola2017image} and diffusion
distillation~\cite{sauer2023adversarial}.
By combining these two approaches, we aim to achieve both few-step sampling and
improved frame appearance control.

For the discriminator $D(\cdot)$ used in the adversarial learning, we adopt a
pixel-space discriminator following previous
methods~\cite{sauer2023stylegan,sauer2023adversarial}.
However, densely applying the adversarial objective to all frames is
computationally expensive, and can lead to degraded temporal
smoothness~\cite{yuan2023instructvideo}.
To mitigate this, drawing inspiration from sparse sampling-based action
recognition~\cite{simonyan2014two,wang2016temporal,wang2017untrimmednets}, we
apply the adversarial loss only to a set of sparsely sampled frames.

Specifically, let $\hat{\bm{x}}_0^{\theta} = \bm{f}_{\theta} (\bm{x}_t, t)$
represent the LCM-predicted video latent, and $\hat{\bm{v}}_0^{\theta} =
\mathscr{D}(\hat{\bm{x}}_0^{\theta})$ be the decoded video, where $\mathscr{D}$
is the latent decoder.
We randomly sample $l$ frames from the video, denoted as
$\{\hat{v}_{0,0}^{\theta,\text{s}}, ...,\hat{v}_{0,l}^{\theta,\text{s}} \}$, and
apply the adversarial loss to these sampled frames:
\begin{equation}
\mathcal{L}_{\text{adv}}^\text{G} = -\mathbb{E}_{\bm{x}_0, \bm{\epsilon}, t}
\left[ \frac{1}{l} \sum_i D_{\psi}(\hat{v}_{0, i}^{\theta,\text{s}}) \right],
\end{equation}
where $\psi$ is the learnable parameter in $D(\cdot)$.
The discriminator is trained to minimize
\begin{equation}
\mathcal{L}_{\text{adv}}^\text{D} = \mathbb{E}_{\bm{x}_0, \bm{\epsilon}, t}
\left[ \max \left( 0, 1 + \frac{1}{l} \sum_i D_{\psi}
(\hat{v}_{0,i}^{\theta,\text{s}}) \right) \right] + \mathbb{E}_{m_i} \left[
\max(0, 1 - D_{\psi} \left( m_i \right) ) \right].
\end{equation}
We omit the gradient penalty
term~\cite{gulrajani2017improved,sauer2023adversarial} for conciseness.
Please find details on the discriminator design
in~\cite{sauer2023stylegan,sauer2023adversarial} or in~\cref{app-sec:additional-details}.
Note that the discriminator's objective is to distinguish between the generated
video frame and the images from the image dataset $\mathbb{M}$, thereby aligning
the video frame with the high-quality distribution.

The overall baseline learning objective is a weighted sum of the CD loss (\cref{eq:cd}) and the
adversarial loss: $\mathcal{L}_{\text{base}} = \mathcal{L}_{\text{CD}} +
\lambda_{\text{adv}} \mathcal{L}_{\text{adv}}^{\text{G}}$, where
$\lambda_{\text{adv}}$ is the weighting hyperparameter.

\begin{figure}[t]
    \centering
    \includegraphics[width=\textwidth]{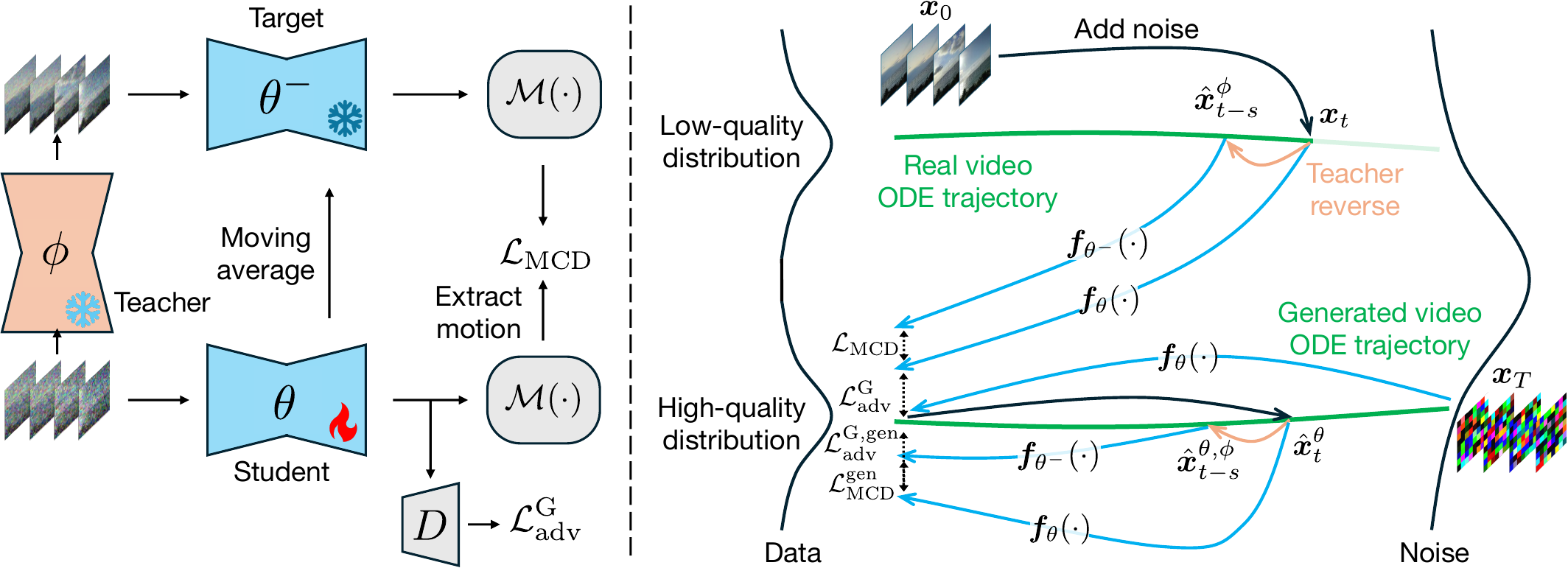}
    \caption{\textbf{Left}: framework overview. Our \methodabbr features disentangled motion-appearance distillation, where motion is learned via the motion consistency distillation loss $\mathcal{L}_{\text{MCD}}$, and the appearance is learned with the frame adversarial objective $\mathcal{L}_{\text{adv}}^{\text{G}}$. \textbf{Right}: mixed trajectory distillation. We simulate the inference-time ODE trajectory using student-generated video (bottom \textcolor{ode_green}{green} line), which is mixed with the real video ODE trajectory (top \textcolor{ode_green}{green} line) for consistency distillation training.}
    \label{fig:framework}
\end{figure}

\subsection{Disentangled motion consistency distillation}

\begin{wrapfigure}{r}{3.2cm}
    \centering
    \includegraphics[width=\linewidth]{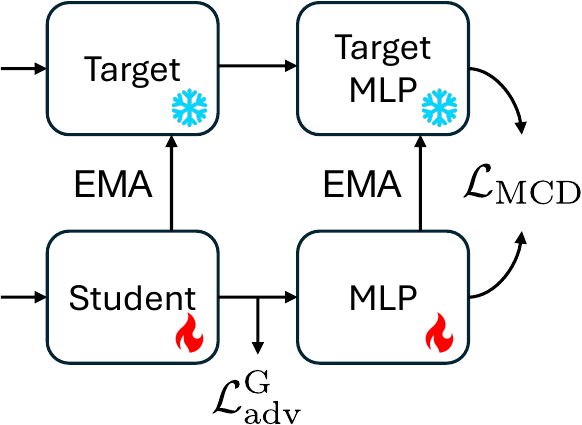}
    \caption{Learnable motion representation.}
    \label{fig:learnable-motion-representation}
\end{wrapfigure}

While combining LCM and the adversarial loss is straightforward, their training
objectives conflict.
The LCM aligns the appearance of the predicted video with the low-quality
distribution, while the adversarial objective adjust the appearance to match the
high-quality distribution.
Moreover, as LCM relies solely on boundary conditions (\cref{sec:preliminaries})
to generate clean video frames, it typically leads to blurry frame as $t$
approaches $T$~\cite{song2023consistency,luo2023latent,wang2024animatelcm}.
This blurriness conflicts with the adversarial loss objective, which seeks to
produce clean frames at all steps.

To mitigate these conflicts, we propose a disentangled approach: separating the
motion presentation from the video latent, and applying the LCM objective
exclusively to the motion representation to ensure temporal smoothness.
Meanwhile, we use the adversarial objective to learn the frame appearance.
This disentanglement allows us to leverage the strengths of both objectives
without them interfering with each other.
An illustration is shown in~\cref{fig:framework} left.

Specifically, we replace the original consistency distillation loss
$\mathcal{L}_{\text{CD}}$ in $\mathcal{L}_{\text{base}}$ with the following motion consistency
distillation (MCD) objective $\mathcal{L}_{\text{MCD}}$:
\begin{equation}
\mathcal{L}_{\text{MCD}} = \mathbb{E}_{\bm{x}_0,\bm{\epsilon},t} \left[ d \left(
\mathcal{M}\left( \bm{f}_{\theta} (\bm{x}_{t}, t)\right), \mathcal{M}\left(
\bm{f}_{\theta^-} (\hat{\bm{x}}_{t-s}^{\phi}, t-s)\right) \right) \right],
\end{equation}
where $\mathcal{M}(\cdot)$ indicates motion representation extraction function.
In this paper, we consider the following implementations.
\begin{itemize}[leftmargin=*,noitemsep,topsep=0pt]
\item \textbf{Latent difference:} the difference between temporally consecutive
frame latents.
This was used in action
recognition~\cite{simonyan2014two,wang2016temporal,zhao2018recognize} as motion representations.
\item \textbf{Latent correlation:} a 4D cost volume tensor containing pairwise
dot-product between consecutive frame latents, widely used in optical flow
estimation~\cite{huang2022flowformer,shi2023flowformer++,shi2023videoflow}.
\item \textbf{Latent low/high-frequency components:}
this is inspired by recent work~\cite{wu2023freeinit}, which shows different
frequency components of the latent have different impacts on the motion.
\item \textbf{Learnable representation:}
we apply a two-layer MLP after the LCM output to extract motion 
(\cref{fig:learnable-motion-representation}).
This MLP does not contain temporal layers, and preserves the motion while
avoiding direct applying appearance loss to the latent.
The parameters of the target MLP head are updated using exponential moving
average, and the MLP is discarded during inference.
This design resembles the projection head used in self-supervised
learning~\cite{grill2020bootstrap,chen2020simple,chen2020improved,caron2021emerging}.
\end{itemize}

We examine all choices in~\cref{subsec:ablation}, and find that the learnable
representation works the best, enabling effective motion consistency modeling
while preserving frame appearance quality.

\subsection{Mixed trajectory distillation}

While the disentangled motion distillation mitigates the appearance learning
objective conflicts, another training-inference discrepancy persists.
During training, all ODE trajectories used for MCD are sampled by adding noise
to the low-quality videos in $\mathbb{V}$ (\cref{fig:framework} top right).
In contrast, during inference, the ODE trajectories will be sampled in the
high-quality video space.
As a result, such discrepancy leads to degraded performance during inference.
Furthermore, since we do not have access to high-quality videos, we cannot directly
sample ODE trajectories from this distribution for training, making it
challenging to resolve this training-inference discrepancy.

To address this problem, we propose to simulate such high-quality video ODE
trajectories by using the consistency model multi-step inference
process~\cite{song2023consistency}.
Specifically, starting from random noise $\bm{x}_{T} \in \mathcal{N} (0, \bm{I})$, we first run single-step
inference to get the high-quality data sample $\hat{\bm{x}}_0^{\theta}$, and
then add noise to it to sample high-quality ODE trajectories.
An illustration is shown in~\cref{fig:framework} bottom right.
Formally, the starting point $\hat{\bm{x}}_t^{\theta}$ for MCD is sampled as:
\begin{equation}
\hat{\bm{x}}_t^{\theta} = \alpha_t \hat{\bm{x}}_0^{\theta} + \sigma_t
\bm{\epsilon}, \ \text{where} \ \hat{\bm{x}}_0^{\theta} =
\bm{f}_{\theta}(\bm{x}_T, T) \ \text{and} \ \bm{x}_T \in \mathcal{N} (0,
\bm{I}).
\end{equation}

In this way, the MCD loss can be formulated by replacing the starting data
points from those sampled from low-quality trajectories with those sampled from
simulated high-quality video trajectories:
\begin{equation}
\mathcal{L}_{\text{MCD}}^{\text{gen}} = \mathbb{E}_{\bm{x}_T, \bm{\epsilon}, t}
\left[ d \left( \mathcal{M}\left( \bm{f}_{\theta} (\hat{\bm{x}}_t^{\theta},
t)\right), \mathcal{M}\left( \bm{f}_{\theta^-}
(\hat{\bm{x}}_{t-s}^{\theta,\phi}, t-s)\right) \right) \right],
\end{equation}
where $\hat{\bm{x}}_{t-s}^{\theta,\phi} = \hat{\bm{x}}_t^{\theta} - s
\bm{\Phi}(\hat{\bm{x}}_t^{\theta}, t; \phi)$.
Similarly, the adversarial loss is also applied to the model prediction from the
samples on the high-quality trajectories, which we denote as
$\mathcal{L}_{\text{adv}}^{\text{G},\text{gen}}$.
By applying the MCD and adversarial objectives to the simulated high-quality
trajectory, it not only mitigates the training-inference trajectory discrepancy,
but also ensures zero terminal signal-to-noise ratio $\bm{x}_T \in \mathcal{N}
(0, \bm{I})$ to avoid the data leakage problem~\cite{lin2024common} and helps plain video distillation.

In practice, we found that only training from the generated video ODE trajectory
potentially leads to mode collapse (\cref{app-subsec:mixed-trajectory-distillation}), and mixing the training from real samples and
generated samples achieves the best results.
Thus, the final loss $\mathcal{L}$ is formulated as a weighted sum of the distillation
objectives using real samples and generated samples:
\begin{equation}
\mathcal{L} = \lambda_{\text{real}} \left(\mathcal{L}_{\text{MCD}} +
\lambda_{\text{adv}} \mathcal{L}_{\text{adv}}^\text{G} \right) + (1 -
\lambda_{\text{real}}) \left( \mathcal{L}_{\text{MCD}}^{\text{gen}} +
\lambda_{\text{adv}} \mathcal{L}_{\text{adv}}^{\text{G},\text{gen}} \right),
\end{equation}
where $\lambda_{\text{real}} \in [0,1]$ is a hyperparameter to balance the
distillation in different trajectories.

\begin{table}[t]
\caption{Video diffusion distillation comparison on the WebVid mini validation set.}
\label{tab:comp-sota-webvid}
\centering
\resizebox{\linewidth}{!}{%
\begin{tabular}{c|c|cccc|cccc}
    \hline 
    \multirow{2}{*}{Teacher} & \multirow{2}{*}{Method} & \multicolumn{4}{c|}{FVD@Step $\downarrow$} & \multicolumn{4}{c}{CLIPSIM@Step $\uparrow$}\tabularnewline
     &  & 1 & 2 & 4 & 8 & 1 & 2 & 4 & 8\tabularnewline
    \hline 
    \multirow{6}{*}{\makecell{AnimateDiff~\cite{guo2023animatediff} \\ $512\times 512\times 16$}} & DDIM~\cite{song2020denoising} & 5228 & 2580 & 1222 & 858 & 20.05 & 20.61 & 24.04 & 28.89\tabularnewline
     & DPM++~\cite{lu2022dpm} & 2082 & 1142 & 843 & 975 & 22.13 & 24.78 & 29.38 & \textbf{30.75}\tabularnewline
     & LCM~\cite{luo2023latent} (our impl.) & 1242 & 978 & 1006 & 909 & 27.32 & 28.95 & 29.94 & 29.86\tabularnewline
     & AnimateLCM~\cite{wang2024animatelcm} & 1575 & 1333 & 998 & 946 & 24.65 & 27.43 & 28.98 & 28.77\tabularnewline
     & AnimateDiff-Lightning~\cite{lin2024animatediff} & 1288 & 1289 & 1283 & 1355 & 28.29 & 29.07 & 30.01 & 29.69\tabularnewline
     & \cellcolor{mygray}\methodabbr (ours) & \cellcolor{mygray}\textbf{1025} & \cellcolor{mygray}\textbf{948} & \cellcolor{mygray}\textbf{827} & \cellcolor{mygray}\textbf{821} & \cellcolor{mygray}\textbf{30.36} & \cellcolor{mygray}\textbf{30.70} & \cellcolor{mygray}\textbf{30.85} & \cellcolor{mygray}29.88\tabularnewline
    \hline 
    \multirow{4}{*}{\makecell{ModelScopeT2V~\cite{wang2023modelscope} \\ $256\times 256\times 16$}} & DDIM~\cite{song2020denoising} & 7231 & 2309 & 1229 & 652 & 20.47 & 20.06 & 23.68 & 28.46\tabularnewline
     & DPM++~\cite{lu2022dpm} & 2030 & 1142 & 477 & 506 & 22.38 & 24.79 & 29.70 & \textbf{31.00}\tabularnewline
     & LCM~\cite{luo2023latent} (our impl.) & 854 & 658 & 603 & 637 & 27.50 & 29.28 & 30.48 & 30.73\tabularnewline
     & \cellcolor{mygray}\methodabbr (ours) & \cellcolor{mygray}\textbf{526} & \cellcolor{mygray}\textbf{450} & \cellcolor{mygray}\textbf{456} & \cellcolor{mygray}\textbf{503} & \cellcolor{mygray}\textbf{29.81} & \cellcolor{mygray}\textbf{30.63} & \cellcolor{mygray}\textbf{30.55} & \cellcolor{mygray}29.58\tabularnewline
    \hline 
\end{tabular}%
}
\end{table}

\section{Experiments}
\label{sec:experiments}

\noindent\textbf{Implementation details.}
We choose two text-to-video diffusion models for experiments:
ModelScopeT2V~\cite{wang2023modelscope} and
AnimateDiff~\cite{guo2023animatediff} with StableDiffusion
v1.5~\cite{rombach2022high}.
We use DDIM solver~\cite{song2020denoising} with 50 sampling steps as the
ODE solver $\bm{\Phi}$.
We train LoRA~\cite{hu2021lora} with rank $64$ as \methodabbr,
following~\cite{luo2023latent,wang2024animatelcm}.
Following the teacher model setting, we generate $16$-frame videos, with
resolution 256x256 for ModelScopeT2V, and 512x512 for AnimateDiff.
The learning rates for the diffusion model and discriminator are set to
$5e^{-6}$ and $5e^{-5}$, respectively, with batch size $128$, Adam
optimizer~\cite{kingma2014adam}, and $30k$ training steps.
The weight hyperparameters are determined via a grid search:
$\lambda_{\text{adv}}=1$ and $\lambda_{\text{real}}=0.5$.
The experiments are conducted on a machine equipped with 32 H100 GPUs.
The project is developed using PyTorch~\cite{ansel2024pytorch},
Diffusers~\cite{von-platen-etal-2022-diffusers}, and PEFT~\cite{peft}.

\noindent\textbf{Evaluation metrics.}
We use FVD~\cite{unterthiner2018towards} to
measure the video quality, and CLIP~\cite{radford2021learning} similarity score
(CLIPSIM) for video-prompt alignment measurement, where CLIP-ViT-B/16 is used.
In~\cref{sec:exp_frame}, we use
FID~\cite{heusel2017gans}
to measure the frame quality and the similarity to the image dataset.

\begin{table}[t]
\caption{Zero-shot video diffusion distillation comparison on the MSRVTT validation set.}
\label{tab:comp-sota-msrvtt}
\centering
\resizebox{\linewidth}{!}{%
\begin{tabular}{c|c|cccc|cccc}
    \hline 
    \multirow{2}{*}{Teacher} & \multirow{2}{*}{Method} & \multicolumn{4}{c|}{FVD@Step $\downarrow$} & \multicolumn{4}{c}{CLIPSIM@Step $\uparrow$}\tabularnewline
     &  & 1 & 2 & 4 & 8 & 1 & 2 & 4 & 8\tabularnewline
    \hline 
    \multirow{6}{*}{\makecell{AnimateDiff~\cite{guo2023animatediff} \\ $512\times 512\times 16$}} & DDIM~\cite{song2020denoising} & 4782 & 4350 & 2774 & 933 & 20.90 & 20.94 & 22.87 & 27.36\tabularnewline
     & DPM++~\cite{lu2022dpm} & 2004 & 1447 & 876 & 794 & 22.93 & 24.5 & 27.62 & 29.10\tabularnewline
     & LCM~\cite{luo2023latent} (our impl.) & 1276 & 1180 & 956 & 830 & 25.75 & 27.33 & 28.37 & 28.65\tabularnewline
     & AnimateLCM~\cite{wang2024animatelcm} & 1578 & 1278 & 824 & 740 & 27.56 & 28.52 & 29.58 & 27.67\tabularnewline
     & AnimateDiff-Lightning~\cite{lin2024animatediff} & 1260 & 1259 & 892 & 932 & 27.38 & 28.77 & 29.12 & 28.77\tabularnewline
     & \cellcolor{mygray}\methodabbr (ours) & \cellcolor{mygray}\textbf{1197} & \cellcolor{mygray}\textbf{1036} & \cellcolor{mygray}\textbf{801} & \cellcolor{mygray}\textbf{675} & \cellcolor{mygray}\textbf{28.95} & \cellcolor{mygray}\textbf{29.40} & \cellcolor{mygray}\textbf{29.64} & \cellcolor{mygray}\textbf{29.13}\tabularnewline
    \hline 
    \multirow{4}{*}{\makecell{ModelScopeT2V~\cite{wang2023modelscope} \\ $256\times 256\times 16$}} & DDIM~\cite{song2020denoising} & 6459 & 2305 & 1445 & 841 & 21.49 & 20.33 & 22.57 & 26.76\tabularnewline
     & DPM++~\cite{lu2022dpm} & 2039 & 1336 & 467 & 552 & 23.48 & 24.85 & 28.51 & \textbf{29.70}\tabularnewline
     & LCM~\cite{luo2023latent} (our impl.) & 1094 & 820 & 713 & 717 & 26.78 & 28.01 & 28.45 & 29.01\tabularnewline
     & \cellcolor{mygray}\methodabbr (ours) & \cellcolor{mygray}\textbf{501} & \cellcolor{mygray}\textbf{434} & \cellcolor{mygray}\textbf{414} & \cellcolor{mygray}\textbf{482} & \cellcolor{mygray}\textbf{28.37} & \cellcolor{mygray}\textbf{29.02} & \cellcolor{mygray}\textbf{28.86} & \cellcolor{mygray}28.28\tabularnewline
    \hline 
\end{tabular}%
}
\end{table}

\begin{figure}[t]
    \centering
    \includegraphics[width=\textwidth]{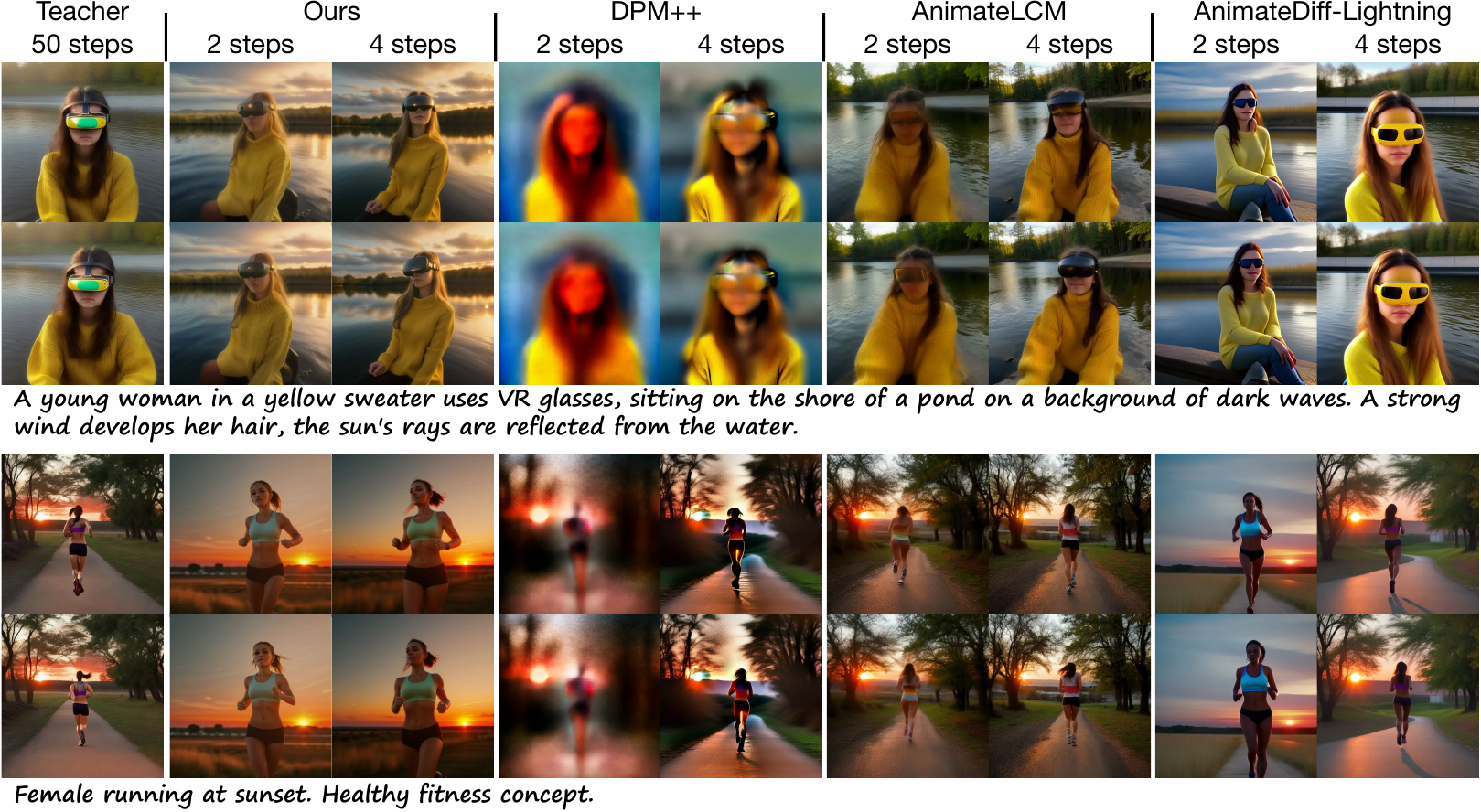}
    \caption{Qualitative comparison of video diffusion distillation with
    AnimateDiff~\cite{guo2023animatediff} as the teacher model. The first and
    last frames are sampled for visualization. Our \methodabbr produces cleaner
    frames using only 2 and 4 sampling steps, with better prompt alignment and
    improved frame details. Corresponding video results are in supplementary
    materials.}
    \label{fig:distillation-comparison}
\end{figure}

\begin{figure}[t]
    \centering
    \includegraphics[width=\textwidth]{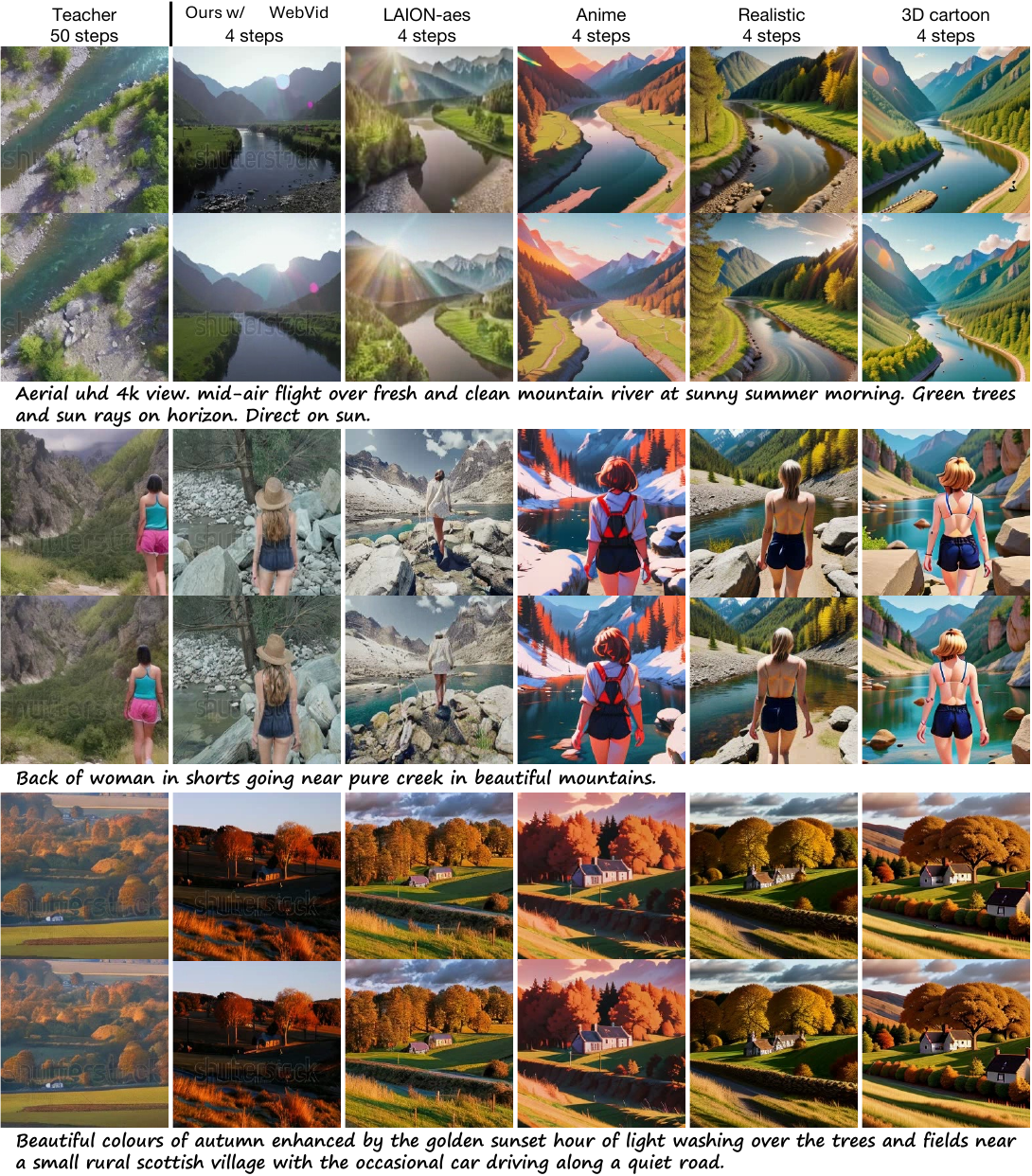}
    \caption{\methodabbr frame quality improvement results using different image
    datasets with ModelScopeT2V~\cite{wang2023modelscope} teacher. The first and
    last frames are sampled for visualization. Our \methodabbr effectively
    adapts to different distributions with 4 steps. Corresponding video
    results are in supplementary materials.}
    \label{fig:appearance-adjustment}
\end{figure}

\begin{table}[t]
\begin{minipage}[h]{0.37\linewidth}
\centering
\caption{Comparison on frame quality improvement. Performance is measured by
FID using different numbers of sampling steps.}
\label{tab:frame-quality-improvement}
\resizebox{\linewidth}{!}{%
\begin{tabular}{c|cccc|cccc}
    \hline 
    \multirow{2}{*}{Method} & \multicolumn{4}{c|}{FID $\downarrow$ (LAION-aes)} & \multicolumn{4}{c}{FID $\downarrow$ (Anime)}\tabularnewline
     & 1 & 2 & 4 & 8 & 1 & 2 & 4 & 8\tabularnewline
    \hline 
    Two-stage & 110 & 119 & 115 & 122 & 160 & 159 & 149 & 150\tabularnewline
    \rowcolor{mygray}\methodabbr (ours) & \textbf{82} & \textbf{69} & \textbf{73} & \textbf{76} & \textbf{78} & \textbf{71} & \textbf{67} & \textbf{69}\tabularnewline
    \hline 
\end{tabular}%
}
\end{minipage}
\hfill
\begin{minipage}[h]{0.6\linewidth}  
\centering
\caption{Ablation study on each components in \methodabbr.
ModelScopeT2V~\cite{wang2023modelscope} is used as teacher.}
\label{tab:main-ablation-study}
\resizebox{\linewidth}{!}{%
\begin{tabular}{ccc|cccc|cccc}
    \hline 
    \multirow{2}{*}{Adv.} & \multirow{2}{*}{MCD} & \multirow{2}{*}{Mixed} & \multicolumn{4}{c|}{FVD@Step $\downarrow$} & \multicolumn{4}{c}{CLIPSIM@Step $\uparrow$}\tabularnewline
     &  &  & 1 & 2 & 4 & 8 & 1 & 2 & 4 & 8\tabularnewline
    \hline 
    \multicolumn{3}{c|}{\rule[3pt]{0.75cm}{0.5pt} LCM~\cite{luo2023latent} \rule[3pt]{0.75cm}{0.5pt}} & 854 & 658 & 603 & 637 & 27.50 & 29.28 & 30.48 & 30.73\tabularnewline
    \rowcolor{grad1}$\checkmark$ &  &  & 703 & 650 & 767 & 760 & 28.70 & 30.37 & \textbf{30.90} & \textbf{30.77}\tabularnewline
    \rowcolor{grad2}$\checkmark$ & $\checkmark$ &  & 563 & 526 & 548 & 579 & 29.30 & 30.40 & 30.11 & 29.52\tabularnewline
    \rowcolor{grad3}$\checkmark$ & $\checkmark$ & $\checkmark$ & \textbf{526} & \textbf{450} & \textbf{456} & \textbf{503} & \textbf{29.81} & \textbf{30.63} & 30.55 & 29.58\tabularnewline
    \hline 
\end{tabular}%
}
\end{minipage}
\end{table}

\begin{table}[t]
\begin{minipage}[h]{0.52\linewidth}
\centering
\caption{Ablation study on motion representation.}
\label{tab:ablation-motion-representation}
\resizebox{\linewidth}{!}{%
\begin{tabular}{c|cccc|cccc}
    \hline 
    \multirow{2}{*}{Motion} & \multicolumn{4}{c|}{FVD@Step $\downarrow$} & \multicolumn{4}{c}{CLIPSIM@Step $\uparrow$}\tabularnewline
     & 1 & 2 & 4 & 8 & 1 & 2 & 4 & 8\tabularnewline
    \hline 
    \rowcolor{grad1}Raw latent & 703 & 650 & 767 & 760 & 28.70 & 30.37 & 30.90 & \textbf{30.77}\tabularnewline
    Diff. & 628 & 640 & 722 & 779 & 29.17 & 30.39 & \textbf{31.03} & 30.55\tabularnewline
    Corr. & 659 & 699 & 727 & 725 & 28.83 & 30.39 & 30.62 & 30.62\tabularnewline
    Low freq. & 677 & 667 & 753 & 919 & 29.25 & 30.32 & 30.35 & 29.79\tabularnewline
    High freq. & 676 & 647 & 687 & 753 & 28.76 & 30.01 & 30.16 & 29.43\tabularnewline
    \rowcolor{grad2}Learnable & \textbf{563} & \textbf{526} & \textbf{548} & \textbf{579} & \textbf{29.30} & \textbf{30.40} & 30.11 & 29.52\tabularnewline
    \hline 
\end{tabular}%
}
\end{minipage}
\hfill
\begin{minipage}[h]{0.47\linewidth}  
\centering
\caption{Ablation study on the mixed trajectory training.}
\label{tab:ablation-mixed-trajectory}
\resizebox{\linewidth}{!}{%
\begin{tabular}{c|cccc|cccc}
    \hline 
    \multirow{2}{*}{$\lambda_{\text{real}}$} & \multicolumn{4}{c|}{FVD@Step $\downarrow$} & \multicolumn{4}{c}{CLIPSIM@Step $\uparrow$}\tabularnewline
     & 1 & 2 & 4 & 8 & 1 & 2 & 4 & 8\tabularnewline
    \hline 
    \rowcolor{grad2}1 & 563 & 526 & 548 & 579 & 29.30 & 30.40 & 30.11 & 29.52\tabularnewline
    \rowcolor{grad3}0.5 & \textbf{526} & \textbf{450} & \textbf{456} & \textbf{503} & \textbf{29.81} & \textbf{30.63} & \textbf{30.55} & \textbf{29.58}\tabularnewline
    0 & 556 & 498 & 483 & 523 & 28.77 & 29.73 & 29.93 & 29.53\tabularnewline
    \hline 
\end{tabular}%
}
\end{minipage}
\end{table}

\subsection{Comparisons with distilled video diffusion models}
\label{sec:exp_video}

\noindent\textbf{Experiment settings.}
In this section, we compare \methodabbr with other video diffusion models to
validate the effectiveness of the disentangled motion-appearance distillation.
For fair comparison, we use the WebVid 2M~\cite{bain2021frozen} as both the
video and image training dataset, without using any additional image datasets.
We postpone the experiments with different image datasets in
Sec.~\ref{sec:exp_frame}.
For testing, we randomly sample 500 validation videos from WebVid 2M (WebVid
mini) for in-distribution evaluation; we also follow common
practice~\cite{wang2023modelscope,ge2023preserve} to use $\sim$2900 validation
videos from MSRVTT~\cite{xu2016msr} for zero-shot generation evaluation. We
report standard metrics of FVD and CLIPSIM.

\begin{figure}[t]
    \centering
    \includegraphics[width=\textwidth]{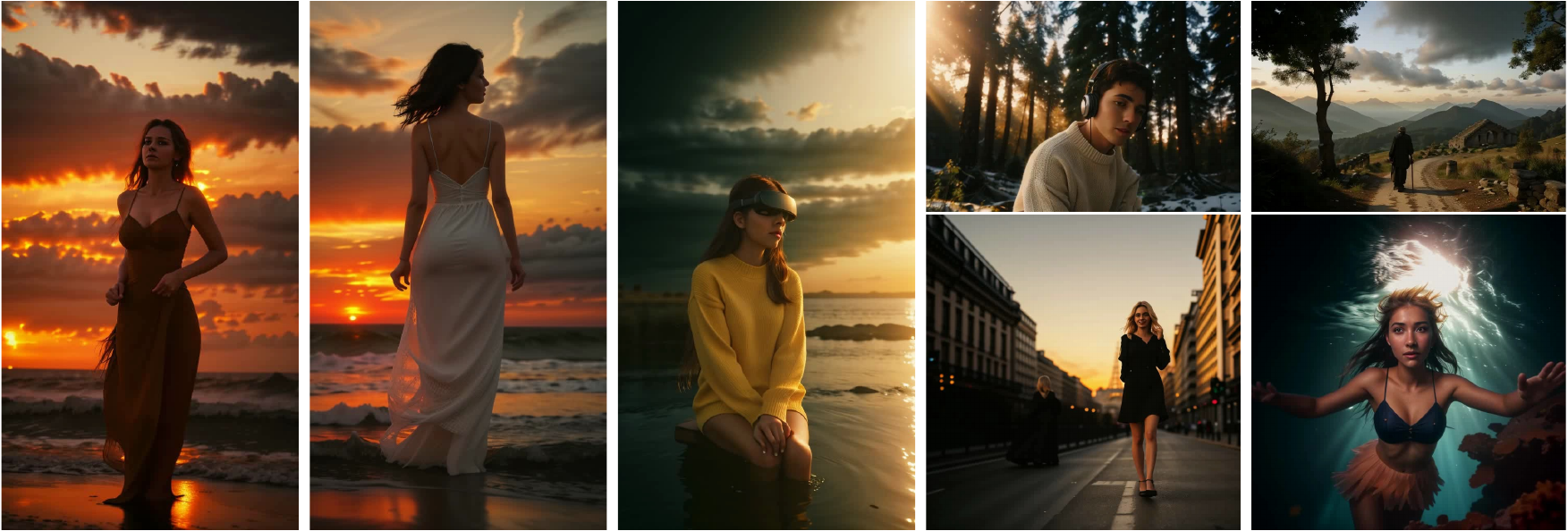}
    \caption{Our \methodabbr allows for high-resolution video generations with
    different aspect ratios, using 4 sampling steps. The left three videos (576x1024) are in portrait format, the top right two videos (768x512) are in landscape format, and the bottom right two videos (768x768) are in square format.}
    \label{fig:aspect-ratio}
\end{figure}

\begin{figure}[t]
    \centering
    \includegraphics[width=\textwidth]{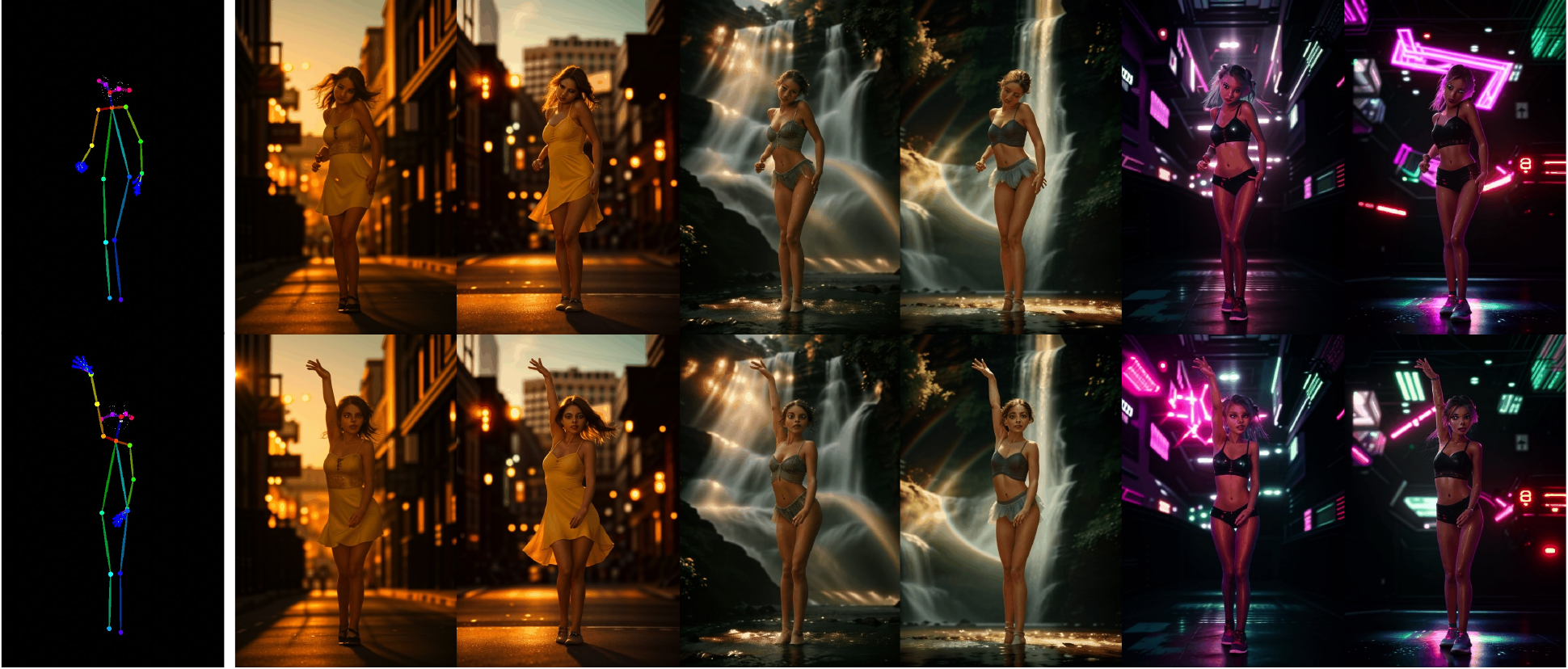}
    \caption{Our \methodabbr can incorporate ControlNet~\cite{zhang2023adding}
    to enable pose-conditioned video generation. The videos are generated using
    4 sampling steps.}
    \label{fig:pose-condition}
\end{figure}

\noindent\textbf{Results.}
We compare our method with training-free samplers DDIM~\cite{song2020denoising}
and DPM++~\cite{lu2022dpm}, and trainable models LCM~\cite{luo2023latent},
AnimateLCM~\cite{wang2024animatelcm} and
AnimateDiff-Lightning~\cite{lin2024animatediff}.
We present the quantitative results on WebVid and MSRVTT in
\cref{tab:comp-sota-webvid} and \cref{tab:comp-sota-msrvtt}, respectively.
On both datasets, and using both teacher models
AnimateDiff~\cite{guo2023animatediff} and
ModelScopeT2V~\cite{wang2023modelscope}, our \methodabbr outperforms all
competing distillation methods in terms of video quality and video-prompt
alignment at sampling steps 1, 2, and 4.
We compare the qualitative results in~\cref{fig:distillation-comparison}, where
our method outputs clean video frames using 2 and 4 sampling steps.
Compared with AnimateDiff-Lightning~\cite{lin2024animatediff}, we achieve better
prompt-alignment and frame details, demonstrating our effectiveness.

\subsection{Video distillation with improved frame quality}
\label{sec:exp_frame}

\noindent\textbf{Experiment settings.}
In this section, we unleash the full capability of \methodabbr by equipping the
model with diverse types of image datasets. We showcase the effectiveness of
using image datasets to boost frame quality as well as adapting the model to new
styles without requiring corresponding video data.
Specifically, we continue to use WebVid 2M~\cite{bain2021frozen} as the video
dataset.
For image dataset, we consider the following options.
LAION-aes, a 4M subset sampled from the high-aesthetic image dataset
LAION-Aesthetic 6+~\cite{schuhmann2022laion}; datasets in anime, realistic, and
3D cartoon styles generated by fine-tuned
StableDiffusion~\cite{toonyou2023,realisticvisionv60b1,disneypixarcartoonb},
each dataset containing 500k images with LAION captions.

\noindent\textbf{Results.}
Due to the lack of prior arts with comparable capabilities, we validate by
comparing \methodabbr with a designed two-stage baseline.
Following AnimateLCM~\cite{wang2024animatelcm}, we implement a two-stage method:
first training the spatial layers with the image dataset, then training the
temporal layers with the video dataset.
We use a real image dataset, LAION-aes, and a generated anime-style dataset for
quantitative comparison.
As shown in \cref{tab:frame-quality-improvement}, our \methodabbr achieves lower
FID scores on both datasets, demonstrating its superior adaptability to
different image dataset distributions.
We present our qualitative results in \cref{fig:appearance-adjustment}, where
our method demonstrates high frame quality using only four sampling steps,
achieving sharp frames with various styles.

\subsection{Ablation study}
\label{subsec:ablation}

We conduct the distillation ablation study on WebVid mini~\cite{bain2021frozen}
using ModelScopeT2V~\cite{wang2023modelscope} as teacher.
The main results are summarized in \cref{tab:main-ablation-study}.
Starting from LCM~\cite{luo2023latent}, we progressively add adversarial
learning, motion consistency distillation loss, and mixed trajectory
distillation.
Each addition shows an improvement in video quality, demonstrating the
effectiveness of these components.

\noindent\textbf{Disentangled motion distribution.}
We compare different motion representations
in~\cref{tab:ablation-motion-representation}.
All representations, except for the latent low-frequency components, help reduce
FVD and improve CLIPSIM at low sampling steps.
This indicates that disentangling motion from the raw latent space for
consistency learning enhances video quality.
Among these, the learnable motion representation performs the best,
demonstrating its effectiveness compared to handcrafted motion representations.

\noindent\textbf{Mixed trajectory distribution.}
\cref{tab:ablation-mixed-trajectory} compares the performance between using only
real video ($\lambda_{\text{real}}=1$), only generated video
($\lambda_{\text{real}}=0$), or a mixed of both.
The results reveal that mixing the trajectories achieves the best results.
While using only generated videos also reduces FVD at all steps, it potentially
leads to mode collapse, producing nearly all-black videos
(\cref{app-subsec:mixed-trajectory-distillation}), highlighting the importance
of incorporating real videos.

\subsection{Applications}

We demonstrate that our proposed \methodabbr is versatile and applicable to
various tasks, including high-resolution video generation with different aspect
ratios and conditioned video generation.
In~\cref{fig:aspect-ratio}, \methodabbr generates high-resolution videos up to
576x1024 pixels and supports various aspect ratios such as square, portrait, and
landscape formats.
Moreover, by incorporating controllable modules, \methodabbr enables conditional
video generation.
For example, in~\cref{fig:pose-condition}, integrating
ControlNet~\cite{zhang2023adding} allows for pose-conditioned video generation.
Notably, both videos are generated using only 4 sampling steps, highlighting the
effectiveness and versatility of our \methodabbr.

\section{Conclusion}
\label{sec:conclusion}

We introduce \methodfull (\methodabbr) to tackle the challenge of low frame quality in video diffusion distillation. %
Our method leverages disentangled motion distillation to separate motion
learning from appearance learning, and mixed trajectory distillation to align
training and inference through simulating the inference ODE trajectory.
\methodabbr achieves state-of-the-art video diffusion
distillation results, and effectively improves frame quality when using a
high-quality image dataset.
\textbf{Broader impacts and limitations.}
\methodabbr accelerates and enhances video generation for creative applications. %
However, as a data-driven approach, our model is sensitive to the distribution and diversity of the training data, which influences model's ability to generate varied video frames, potentially limiting generalization ability.
\methodabbr also carries potential risks such as deepfakes creation. 
Implementing responsible deployment %
is essential to mitigate these risks.

\section*{Acknowledgement}
This work is supported in part by the Defense Advanced Research Projects Agency (DARPA) under Contract No.~HR001120C0124. Any opinions, findings and conclusions or recommendations expressed in this material are those of the author(s) and do not necessarily reflect the views of the Defense Advanced Research Projects Agency (DARPA).

\bibliographystyle{plain}
\bibliography{ref}

\begin{thebibliography}{10}

\bibitem{disneypixarcartoonb}
Disney pixar cartoon type b - v1.0.
\newblock \url{https://civitai.com/models/75650/disney-pixar-cartoon-type-b},
  2023.

\bibitem{realisticvisionv60b1}
Realistic vision v6.0 b1.
\newblock \url{https://civitai.com/models/4201/realistic-vision-v60-b1}, 2023.

\bibitem{toonyou2023}
Toonyou - beta 6.
\newblock \url{https://civitai.com/models/30240/toonyou}, 2023.

\bibitem{ansel2024pytorch}
Jason Ansel, Edward Yang, Horace He, Natalia Gimelshein, Animesh Jain, Michael
  Voznesensky, Bin Bao, Peter Bell, David Berard, Evgeni Burovski, et~al.
\newblock Pytorch 2: Faster machine learning through dynamic python bytecode
  transformation and graph compilation.
\newblock In {\em Proceedings of the 29th ACM International Conference on
  Architectural Support for Programming Languages and Operating Systems, Volume
  2}, pages 929--947, 2024.

\bibitem{bain2021frozen}
Max Bain, Arsha Nagrani, G{\"u}l Varol, and Andrew Zisserman.
\newblock Frozen in time: A joint video and image encoder for end-to-end
  retrieval.
\newblock In {\em Int. Conf. Comput. Vis.}, pages 1728--1738, 2021.

\bibitem{balaji2019conditional}
Yogesh Balaji, Martin~Renqiang Min, Bing Bai, Rama Chellappa, and Hans~Peter
  Graf.
\newblock Conditional gan with discriminative filter generation for
  text-to-video synthesis.
\newblock In {\em IJCAI}, volume~1, page~2, 2019.

\bibitem{bao2022analytic}
Fan Bao, Chongxuan Li, Jun Zhu, and Bo~Zhang.
\newblock Analytic-dpm: an analytic estimate of the optimal reverse variance in
  diffusion probabilistic models.
\newblock {\em arXiv preprint arXiv:2201.06503}, 2022.

\bibitem{betker2023improving}
James Betker, Gabriel Goh, Li~Jing, Tim Brooks, Jianfeng Wang, Linjie Li, Long
  Ouyang, Juntang Zhuang, Joyce Lee, Yufei Guo, et~al.
\newblock Improving image generation with better captions.
\newblock {\em Computer Science. https://cdn. openai. com/papers/dall-e-3.
  pdf}, 2(3):8, 2023.

\bibitem{blattmann2023align}
Andreas Blattmann, Robin Rombach, Huan Ling, Tim Dockhorn, Seung~Wook Kim,
  Sanja Fidler, and Karsten Kreis.
\newblock Align your latents: High-resolution video synthesis with latent
  diffusion models.
\newblock In {\em IEEE Conf. Comput. Vis. Pattern Recog.}, pages 22563--22575,
  2023.

\bibitem{caron2021emerging}
Mathilde Caron, Hugo Touvron, Ishan Misra, Herv{\'e} J{\'e}gou, Julien Mairal,
  Piotr Bojanowski, and Armand Joulin.
\newblock Emerging properties in self-supervised vision transformers.
\newblock In {\em Int. Conf. Comput. Vis.}, pages 9650--9660, 2021.

\bibitem{chen2020simple}
Ting Chen, Simon Kornblith, Mohammad Norouzi, and Geoffrey Hinton.
\newblock A simple framework for contrastive learning of visual
  representations.
\newblock pages 1597--1607. PMLR, 2020.

\bibitem{chen2020improved}
Xinlei Chen, Haoqi Fan, Ross Girshick, and Kaiming He.
\newblock Improved baselines with momentum contrastive learning.
\newblock {\em arXiv preprint arXiv:2003.04297}, 2020.

\bibitem{dai2023emu}
Xiaoliang Dai, Ji~Hou, Chih-Yao Ma, Sam Tsai, Jialiang Wang, Rui Wang, Peizhao
  Zhang, Simon Vandenhende, Xiaofang Wang, Abhimanyu Dubey, et~al.
\newblock Emu: Enhancing image generation models using photogenic needles in a
  haystack.
\newblock {\em arXiv preprint arXiv:2309.15807}, 2023.

\bibitem{darcet2023vitneedreg}
Timothée Darcet, Maxime Oquab, Julien Mairal, and Piotr Bojanowski.
\newblock Vision transformers need registers, 2023.

\bibitem{fang2024structural}
Gongfan Fang, Xinyin Ma, and Xinchao Wang.
\newblock Structural pruning for diffusion models.
\newblock {\em Adv. Neural Inform. Process. Syst.}, 36, 2024.

\bibitem{ge2023preserve}
Songwei Ge, Seungjun Nah, Guilin Liu, Tyler Poon, Andrew Tao, Bryan Catanzaro,
  David Jacobs, Jia-Bin Huang, Ming-Yu Liu, and Yogesh Balaji.
\newblock Preserve your own correlation: A noise prior for video diffusion
  models.
\newblock In {\em Int. Conf. Comput. Vis.}, pages 22930--22941, 2023.

\bibitem{grill2020bootstrap}
Jean-Bastien Grill, Florian Strub, Florent Altch{\'e}, Corentin Tallec, Pierre
  Richemond, Elena Buchatskaya, Carl Doersch, Bernardo Avila~Pires, Zhaohan
  Guo, Mohammad Gheshlaghi~Azar, et~al.
\newblock Bootstrap your own latent-a new approach to self-supervised learning.
\newblock {\em Adv. Neural Inform. Process. Syst.}, 33:21271--21284, 2020.

\bibitem{gulrajani2017improved}
Ishaan Gulrajani, Faruk Ahmed, Martin Arjovsky, Vincent Dumoulin, and Aaron~C
  Courville.
\newblock Improved training of wasserstein gans.
\newblock {\em Adv. Neural Inform. Process. Syst.}, 30, 2017.

\bibitem{guo2023animatediff}
Yuwei Guo, Ceyuan Yang, Anyi Rao, Yaohui Wang, Yu~Qiao, Dahua Lin, and Bo~Dai.
\newblock Animatediff: Animate your personalized text-to-image diffusion models
  without specific tuning.
\newblock {\em arXiv preprint arXiv:2307.04725}, 2023.

\bibitem{heusel2017gans}
Martin Heusel, Hubert Ramsauer, Thomas Unterthiner, Bernhard Nessler, and Sepp
  Hochreiter.
\newblock Gans trained by a two time-scale update rule converge to a local nash
  equilibrium.
\newblock {\em Adv. Neural Inform. Process. Syst.}, 30, 2017.

\bibitem{ho2022imagen}
Jonathan Ho, William Chan, Chitwan Saharia, Jay Whang, Ruiqi Gao, Alexey
  Gritsenko, Diederik~P Kingma, Ben Poole, Mohammad Norouzi, David~J Fleet,
  et~al.
\newblock Imagen video: High definition video generation with diffusion models.
\newblock {\em arXiv preprint arXiv:2210.02303}, 2022.

\bibitem{ho2020denoising}
Jonathan Ho, Ajay Jain, and Pieter Abbeel.
\newblock Denoising diffusion probabilistic models.
\newblock {\em Adv. Neural Inform. Process. Syst.}, 33:6840--6851, 2020.

\bibitem{ho2022classifier}
Jonathan Ho and Tim Salimans.
\newblock Classifier-free diffusion guidance.
\newblock {\em arXiv preprint arXiv:2207.12598}, 2022.

\bibitem{ho2022video}
Jonathan Ho, Tim Salimans, Alexey Gritsenko, William Chan, Mohammad Norouzi,
  and David~J Fleet.
\newblock Video diffusion models.
\newblock {\em Adv. Neural Inform. Process. Syst.}, 35:8633--8646, 2022.

\bibitem{hu2021lora}
Edward~J Hu, Yelong Shen, Phillip Wallis, Zeyuan Allen-Zhu, Yuanzhi Li, Shean
  Wang, Lu~Wang, and Weizhu Chen.
\newblock Lora: Low-rank adaptation of large language models.
\newblock {\em arXiv preprint arXiv:2106.09685}, 2021.

\bibitem{huang2022flowformer}
Zhaoyang Huang, Xiaoyu Shi, Chao Zhang, Qiang Wang, Ka~Chun Cheung, Hongwei
  Qin, Jifeng Dai, and Hongsheng Li.
\newblock Flowformer: A transformer architecture for optical flow.
\newblock In {\em Eur. Conf. Comput. Vis.}, pages 668--685. Springer, 2022.

\bibitem{isola2017image}
Phillip Isola, Jun-Yan Zhu, Tinghui Zhou, and Alexei~A Efros.
\newblock Image-to-image translation with conditional adversarial networks.
\newblock In {\em IEEE Conf. Comput. Vis. Pattern Recog.}, pages 1125--1134,
  2017.

\bibitem{karras2022elucidating}
Tero Karras, Miika Aittala, Timo Aila, and Samuli Laine.
\newblock Elucidating the design space of diffusion-based generative models.
\newblock {\em Adv. Neural Inform. Process. Syst.}, 35:26565--26577, 2022.

\bibitem{khachatryan2023text2video}
Levon Khachatryan, Andranik Movsisyan, Vahram Tadevosyan, Roberto Henschel,
  Zhangyang Wang, Shant Navasardyan, and Humphrey Shi.
\newblock Text2video-zero: Text-to-image diffusion models are zero-shot video
  generators.
\newblock In {\em Int. Conf. Comput. Vis.}, pages 15954--15964, 2023.

\bibitem{kingma2014adam}
Diederik~P Kingma and Jimmy Ba.
\newblock Adam: A method for stochastic optimization.
\newblock {\em arXiv preprint arXiv:1412.6980}, 2014.

\bibitem{kohler2024imagine}
Jonas Kohler, Albert Pumarola, Edgar Sch{\"o}nfeld, Artsiom Sanakoyeu, Roshan
  Sumbaly, Peter Vajda, and Ali Thabet.
\newblock Imagine flash: Accelerating emu diffusion models with backward
  distillation.
\newblock {\em arXiv preprint arXiv:2405.05224}, 2024.

\bibitem{li2024snapfusion}
Yanyu Li, Huan Wang, Qing Jin, Ju~Hu, Pavlo Chemerys, Yun Fu, Yanzhi Wang,
  Sergey Tulyakov, and Jian Ren.
\newblock Snapfusion: Text-to-image diffusion model on mobile devices within
  two seconds.
\newblock {\em Adv. Neural Inform. Process. Syst.}, 36, 2024.

\bibitem{lin2024common}
Shanchuan Lin, Bingchen Liu, Jiashi Li, and Xiao Yang.
\newblock Common diffusion noise schedules and sample steps are flawed.
\newblock In {\em IEEE Winter Conf. Appl. Comput. Vis.}, pages 5404--5411,
  2024.

\bibitem{lin2024sdxl}
Shanchuan Lin, Anran Wang, and Xiao Yang.
\newblock Sdxl-lightning: Progressive adversarial diffusion distillation.
\newblock {\em arXiv preprint arXiv:2402.13929}, 2024.

\bibitem{lin2024animatediff}
Shanchuan Lin and Xiao Yang.
\newblock Animatediff-lightning: Cross-model diffusion distillation.
\newblock {\em arXiv preprint arXiv:2403.12706}, 2024.

\bibitem{liu2022pseudo}
Luping Liu, Yi~Ren, Zhijie Lin, and Zhou Zhao.
\newblock Pseudo numerical methods for diffusion models on manifolds.
\newblock {\em arXiv preprint arXiv:2202.09778}, 2022.

\bibitem{lu2022dpm}
Cheng Lu, Yuhao Zhou, Fan Bao, Jianfei Chen, Chongxuan Li, and Jun Zhu.
\newblock Dpm-solver: A fast ode solver for diffusion probabilistic model
  sampling in around 10 steps.
\newblock {\em Adv. Neural Inform. Process. Syst.}, 35:5775--5787, 2022.

\bibitem{luo2023latent}
Simian Luo, Yiqin Tan, Longbo Huang, Jian Li, and Hang Zhao.
\newblock Latent consistency models: Synthesizing high-resolution images with
  few-step inference.
\newblock {\em arXiv preprint arXiv:2310.04378}, 2023.

\bibitem{ma2023deepcache}
Xinyin Ma, Gongfan Fang, and Xinchao Wang.
\newblock Deepcache: Accelerating diffusion models for free.
\newblock {\em arXiv preprint arXiv:2312.00858}, 2023.

\bibitem{peft}
Sourab Mangrulkar, Sylvain Gugger, Lysandre Debut, Younes Belkada, Sayak Paul,
  and Benjamin Bossan.
\newblock Peft: State-of-the-art parameter-efficient fine-tuning methods.
\newblock \url{https://github.com/huggingface/peft}, 2022.

\bibitem{meng2023distillation}
Chenlin Meng, Robin Rombach, Ruiqi Gao, Diederik Kingma, Stefano Ermon,
  Jonathan Ho, and Tim Salimans.
\newblock On distillation of guided diffusion models.
\newblock In {\em IEEE Conf. Comput. Vis. Pattern Recog.}, pages 14297--14306,
  2023.

\bibitem{nichol2021improved}
Alexander~Quinn Nichol and Prafulla Dhariwal.
\newblock Improved denoising diffusion probabilistic models.
\newblock pages 8162--8171. PMLR, 2021.

\bibitem{oquab2023dinov2}
Maxime Oquab, Timothée Darcet, Theo Moutakanni, Huy~V. Vo, Marc Szafraniec,
  Vasil Khalidov, Pierre Fernandez, Daniel Haziza, Francisco Massa, Alaaeldin
  El-Nouby, Russell Howes, Po-Yao Huang, Hu~Xu, Vasu Sharma, Shang-Wen Li,
  Wojciech Galuba, Mike Rabbat, Mido Assran, Nicolas Ballas, Gabriel Synnaeve,
  Ishan Misra, Herve Jegou, Julien Mairal, Patrick Labatut, Armand Joulin, and
  Piotr Bojanowski.
\newblock Dinov2: Learning robust visual features without supervision, 2023.

\bibitem{radford2021learning}
Alec Radford, Jong~Wook Kim, Chris Hallacy, Aditya Ramesh, Gabriel Goh,
  Sandhini Agarwal, Girish Sastry, Amanda Askell, Pamela Mishkin, Jack Clark,
  et~al.
\newblock Learning transferable visual models from natural language
  supervision.
\newblock pages 8748--8763. PMLR, 2021.

\bibitem{rombach2022high}
Robin Rombach, Andreas Blattmann, Dominik Lorenz, Patrick Esser, and Bj{\"o}rn
  Ommer.
\newblock High-resolution image synthesis with latent diffusion models.
\newblock In {\em IEEE Conf. Comput. Vis. Pattern Recog.}, pages 10684--10695,
  2022.

\bibitem{salimans2022progressive}
Tim Salimans and Jonathan Ho.
\newblock Progressive distillation for fast sampling of diffusion models.
\newblock {\em arXiv preprint arXiv:2202.00512}, 2022.

\bibitem{sauer2023stylegan}
Axel Sauer, Tero Karras, Samuli Laine, Andreas Geiger, and Timo Aila.
\newblock Stylegan-t: Unlocking the power of gans for fast large-scale
  text-to-image synthesis.
\newblock pages 30105--30118. PMLR, 2023.

\bibitem{sauer2023adversarial}
Axel Sauer, Dominik Lorenz, Andreas Blattmann, and Robin Rombach.
\newblock Adversarial diffusion distillation.
\newblock {\em arXiv preprint arXiv:2311.17042}, 2023.

\bibitem{schuhmann2022laion}
Christoph Schuhmann, Romain Beaumont, Richard Vencu, Cade Gordon, Ross
  Wightman, Mehdi Cherti, Theo Coombes, Aarush Katta, Clayton Mullis, Mitchell
  Wortsman, et~al.
\newblock Laion-5b: An open large-scale dataset for training next generation
  image-text models.
\newblock {\em Adv. Neural Inform. Process. Syst.}, 35:25278--25294, 2022.

\bibitem{shi2023videoflow}
Xiaoyu Shi, Zhaoyang Huang, Weikang Bian, Dasong Li, Manyuan Zhang, Ka~Chun
  Cheung, Simon See, Hongwei Qin, Jifeng Dai, and Hongsheng Li.
\newblock Videoflow: Exploiting temporal cues for multi-frame optical flow
  estimation.
\newblock In {\em Int. Conf. Comput. Vis.}, pages 12469--12480, 2023.

\bibitem{shi2023flowformer++}
Xiaoyu Shi, Zhaoyang Huang, Dasong Li, Manyuan Zhang, Ka~Chun Cheung, Simon
  See, Hongwei Qin, Jifeng Dai, and Hongsheng Li.
\newblock Flowformer++: Masked cost volume autoencoding for pretraining optical
  flow estimation.
\newblock In {\em IEEE Conf. Comput. Vis. Pattern Recog.}, pages 1599--1610,
  2023.

\bibitem{simonyan2014two}
Karen Simonyan and Andrew Zisserman.
\newblock Two-stream convolutional networks for action recognition in videos.
\newblock {\em Adv. Neural Inform. Process. Syst.}, 27, 2014.

\bibitem{skorokhodov2022stylegan}
Ivan Skorokhodov, Sergey Tulyakov, and Mohamed Elhoseiny.
\newblock Stylegan-v: A continuous video generator with the price, image
  quality and perks of stylegan2.
\newblock In {\em IEEE Conf. Comput. Vis. Pattern Recog.}, pages 3626--3636,
  2022.

\bibitem{song2020denoising}
Jiaming Song, Chenlin Meng, and Stefano Ermon.
\newblock Denoising diffusion implicit models.
\newblock {\em arXiv preprint arXiv:2010.02502}, 2020.

\bibitem{song2023consistency}
Yang Song, Prafulla Dhariwal, Mark Chen, and Ilya Sutskever.
\newblock Consistency models.
\newblock {\em arXiv preprint arXiv:2303.01469}, 2023.

\bibitem{song2020score}
Yang Song, Jascha Sohl-Dickstein, Diederik~P Kingma, Abhishek Kumar, Stefano
  Ermon, and Ben Poole.
\newblock Score-based generative modeling through stochastic differential
  equations.
\newblock {\em arXiv preprint arXiv:2011.13456}, 2020.

\bibitem{soomro2012ucf101}
Khurram Soomro, Amir~Roshan Zamir, and Mubarak Shah.
\newblock Ucf101: A dataset of 101 human actions classes from videos in the
  wild.
\newblock {\em arXiv preprint arXiv:1212.0402}, 2012.

\bibitem{tulyakov2018mocogan}
Sergey Tulyakov, Ming-Yu Liu, Xiaodong Yang, and Jan Kautz.
\newblock Mocogan: Decomposing motion and content for video generation.
\newblock In {\em IEEE Conf. Comput. Vis. Pattern Recog.}, pages 1526--1535,
  2018.

\bibitem{unterthiner2018towards}
Thomas Unterthiner, Sjoerd Van~Steenkiste, Karol Kurach, Raphael Marinier,
  Marcin Michalski, and Sylvain Gelly.
\newblock Towards accurate generative models of video: A new metric \&
  challenges.
\newblock {\em arXiv preprint arXiv:1812.01717}, 2018.

\bibitem{von-platen-etal-2022-diffusers}
Patrick von Platen, Suraj Patil, Anton Lozhkov, Pedro Cuenca, Nathan Lambert,
  Kashif Rasul, Mishig Davaadorj, Dhruv Nair, Sayak Paul, William Berman, Yiyi
  Xu, Steven Liu, and Thomas Wolf.
\newblock Diffusers: State-of-the-art diffusion models.
\newblock \url{https://github.com/huggingface/diffusers}, 2022.

\bibitem{wang2024animatelcm}
Fu-Yun Wang, Zhaoyang Huang, Xiaoyu Shi, Weikang Bian, Guanglu Song, Yu~Liu,
  and Hongsheng Li.
\newblock Animatelcm: Accelerating the animation of personalized diffusion
  models and adapters with decoupled consistency learning.
\newblock {\em arXiv preprint arXiv:2402.00769}, 2024.

\bibitem{wang2023modelscope}
Jiuniu Wang, Hangjie Yuan, Dayou Chen, Yingya Zhang, Xiang Wang, and Shiwei
  Zhang.
\newblock Modelscope text-to-video technical report.
\newblock {\em arXiv preprint arXiv:2308.06571}, 2023.

\bibitem{wang2017untrimmednets}
Limin Wang, Yuanjun Xiong, Dahua Lin, and Luc Van~Gool.
\newblock Untrimmednets for weakly supervised action recognition and detection.
\newblock In {\em IEEE Conf. Comput. Vis. Pattern Recog.}, pages 4325--4334,
  2017.

\bibitem{wang2016temporal}
Limin Wang, Yuanjun Xiong, Zhe Wang, Yu~Qiao, Dahua Lin, Xiaoou Tang, and Luc
  Van~Gool.
\newblock Temporal segment networks: Towards good practices for deep action
  recognition.
\newblock In {\em Eur. Conf. Comput. Vis.}, pages 20--36. Springer, 2016.

\bibitem{Wang_2024_CVPR}
Tan Wang, Linjie Li, Kevin Lin, Yuanhao Zhai, Chung-Ching Lin, Zhengyuan Yang,
  Hanwang Zhang, Zicheng Liu, and Lijuan Wang.
\newblock Disco: Disentangled control for realistic human dance generation.
\newblock In {\em IEEE Conf. Comput. Vis. Pattern Recog.}, pages 9326--9336,
  2024.

\bibitem{wang2024videocomposer}
Xiang Wang, Hangjie Yuan, Shiwei Zhang, Dayou Chen, Jiuniu Wang, Yingya Zhang,
  Yujun Shen, Deli Zhao, and Jingren Zhou.
\newblock Videocomposer: Compositional video synthesis with motion
  controllability.
\newblock {\em Adv. Neural Inform. Process. Syst.}, 36, 2024.

\bibitem{wang2023videolcm}
Xiang Wang, Shiwei Zhang, Han Zhang, Yu~Liu, Yingya Zhang, Changxin Gao, and
  Nong Sang.
\newblock Videolcm: Video latent consistency model.
\newblock {\em arXiv preprint arXiv:2312.09109}, 2023.

\bibitem{wang2020g3an}
Yaohui Wang, Piotr Bilinski, Francois Bremond, and Antitza Dantcheva.
\newblock G3an: Disentangling appearance and motion for video generation.
\newblock In {\em IEEE Conf. Comput. Vis. Pattern Recog.}, pages 5264--5273,
  2020.

\bibitem{wu2023tune}
Jay~Zhangjie Wu, Yixiao Ge, Xintao Wang, Stan~Weixian Lei, Yuchao Gu, Yufei
  Shi, Wynne Hsu, Ying Shan, Xiaohu Qie, and Mike~Zheng Shou.
\newblock Tune-a-video: One-shot tuning of image diffusion models for
  text-to-video generation.
\newblock In {\em Int. Conf. Comput. Vis.}, pages 7623--7633, 2023.

\bibitem{wu2023freeinit}
Tianxing Wu, Chenyang Si, Yuming Jiang, Ziqi Huang, and Ziwei Liu.
\newblock Freeinit: Bridging initialization gap in video diffusion models.
\newblock {\em arXiv preprint arXiv:2312.07537}, 2023.

\bibitem{xu2016msr}
Jun Xu, Tao Mei, Ting Yao, and Yong Rui.
\newblock Msr-vtt: A large video description dataset for bridging video and
  language.
\newblock In {\em IEEE Conf. Comput. Vis. Pattern Recog.}, pages 5288--5296,
  2016.

\bibitem{xu2023ufogen}
Yanwu Xu, Yang Zhao, Zhisheng Xiao, and Tingbo Hou.
\newblock Ufogen: You forward once large scale text-to-image generation via
  diffusion gans.
\newblock {\em arXiv preprint arXiv:2311.09257}, 2023.

\bibitem{yuan2023instructvideo}
Hangjie Yuan, Shiwei Zhang, Xiang Wang, Yujie Wei, Tao Feng, Yining Pan, Yingya
  Zhang, Ziwei Liu, Samuel Albanie, and Dong Ni.
\newblock Instructvideo: Instructing video diffusion models with human
  feedback.
\newblock {\em arXiv preprint arXiv:2312.12490}, 2023.

\bibitem{zhai2024idol}
Yuanhao Zhai, Kevin Lin, Linjie Li, Chung-Ching Lin, Jianfeng Wang, Zhengyuan
  Yang, David Doermann, Junsong Yuan, Zicheng Liu, and Lijuan Wang.
\newblock Idol: Unified dual-modal latent diffusion for human-centric joint
  video-depth generation.
\newblock In {\em Eur. Conf. Comput. Vis.}, 2024.

\bibitem{zhang2023adding}
Lvmin Zhang, Anyi Rao, and Maneesh Agrawala.
\newblock Adding conditional control to text-to-image diffusion models.
\newblock In {\em Int. Conf. Comput. Vis.}, pages 3836--3847, 2023.

\bibitem{zhang2022fast}
Qinsheng Zhang and Yongxin Chen.
\newblock Fast sampling of diffusion models with exponential integrator.
\newblock {\em arXiv preprint arXiv:2204.13902}, 2022.

\bibitem{zhao2018recognize}
Yue Zhao, Yuanjun Xiong, and Dahua Lin.
\newblock Recognize actions by disentangling components of dynamics.
\newblock In {\em IEEE Conf. Comput. Vis. Pattern Recog.}, pages 6566--6575,
  2018.

\bibitem{zhu2017unpaired}
Jun-Yan Zhu, Taesung Park, Phillip Isola, and Alexei~A Efros.
\newblock Unpaired image-to-image translation using cycle-consistent
  adversarial networks.
\newblock In {\em Int. Conf. Comput. Vis.}, pages 2223--2232, 2017.

\end{thebibliography}

\clearpage

\appendix

\section{Related work}
\label{sec:related}

\noindent\textbf{Video generation.}
Early methods mainly rely on generative adversarial networks
(GANs)~\cite{tulyakov2018mocogan,balaji2019conditional,wang2020g3an,skorokhodov2022stylegan},
which struggled with scalability and generalization to out-of-distribution
domains.
Building on the recent success of image diffusion
models~\cite{ho2020denoising,rombach2022high}, several methods leverage
diffusion models for video
generation~\cite{ho2022imagen,ho2022video,blattmann2023align,wang2023modelscope,guo2023animatediff,Wang_2024_CVPR,wang2024videocomposer,zhai2024idol}.
This paper focuses on two popular text-to-video diffusion models:
ModelScopeT2V~\cite{wang2023modelscope} and
AnimateDiff~\cite{guo2023animatediff}.
ModelScopeT2V~\cite{wang2023modelscope} inserts spatio-temporal blocks into 2D
backbones and trains the whole backbone end-to-end, ensuring adaptability to
varying frame lengths with smooth motions.
AnimateDiff~\cite{guo2023animatediff} introduces a plug-and-play motion module
that learns motion priors from real-world videos and can be applied to
personalized image diffusion models for video generation.

\noindent\textbf{Diffusion distillation.}
Training-free methods primarily focus on developing advanced ODE solvers to
reduce the number of sampling
steps~\cite{song2020denoising,lu2022dpm,bao2022analytic,liu2022pseudo,zhang2022fast}.
Another approach employs smaller
backbones~\cite{ma2023deepcache,fang2024structural,li2024snapfusion} to improve
structural efficiency.
To further reduce the inference steps, progressive
distillation~\cite{salimans2022progressive,meng2023distillation,lin2024sdxl,lin2024animatediff}
is first proposed to distill the two or more steps into one.
Consistency
models~\cite{song2023consistency,luo2023latent,wang2023videolcm,wang2024animatelcm}
generalize progressive distillation by mapping any point on the PF-ODE
trajectory to the same origin.
A parallel line of
work~\cite{xu2023ufogen,sauer2023adversarial,lin2024sdxl,lin2024animatediff,kohler2024imagine}
directly applies constraints on the model output by leveraging adversarial
losses.
However, these methods heavily rely on the quality of the training dataset, and
low-quality datasets inevitably degrade the generation quality.

\noindent\textbf{Frame quality adjustment.}
Unlike image generation, which benefits from high-quality
datasets~\cite{schuhmann2022laion}, popular public video
datasets~\cite{soomro2012ucf101,xu2016msr,bain2021frozen} often exhibit lower
frame quality, including issues like motion blur, low resolution, and
watermarks.
This discrepancy hinders the development of high-quality video generation in a
single-stage, end-to-end manner. 
A straightforward way to improve frame quality is two-stage
approach~\cite{blattmann2023align,guo2023animatediff,wang2024animatelcm}, which
involves training an image backbone with a high-quality image dataset and then
adding temporal layers for fine-tuning on video datasets
Alternatively, some methods~\cite{khachatryan2023text2video,wu2023tune} animate
pre-trained image diffusion models with low-shot fine-tuning.
In this paper, we propose a single-stage, end-to-end method to address this
problem, while simultaneously achieving few-step sampling with video diffusion
distillation.

\section{Additional experiments}
\label{app:sec:additional-exp}

\subsection{Mixed trajectory distillation}
\label{app-subsec:mixed-trajectory-distillation}

\begin{figure}[h]
    \centering
    \includegraphics[width=\textwidth]{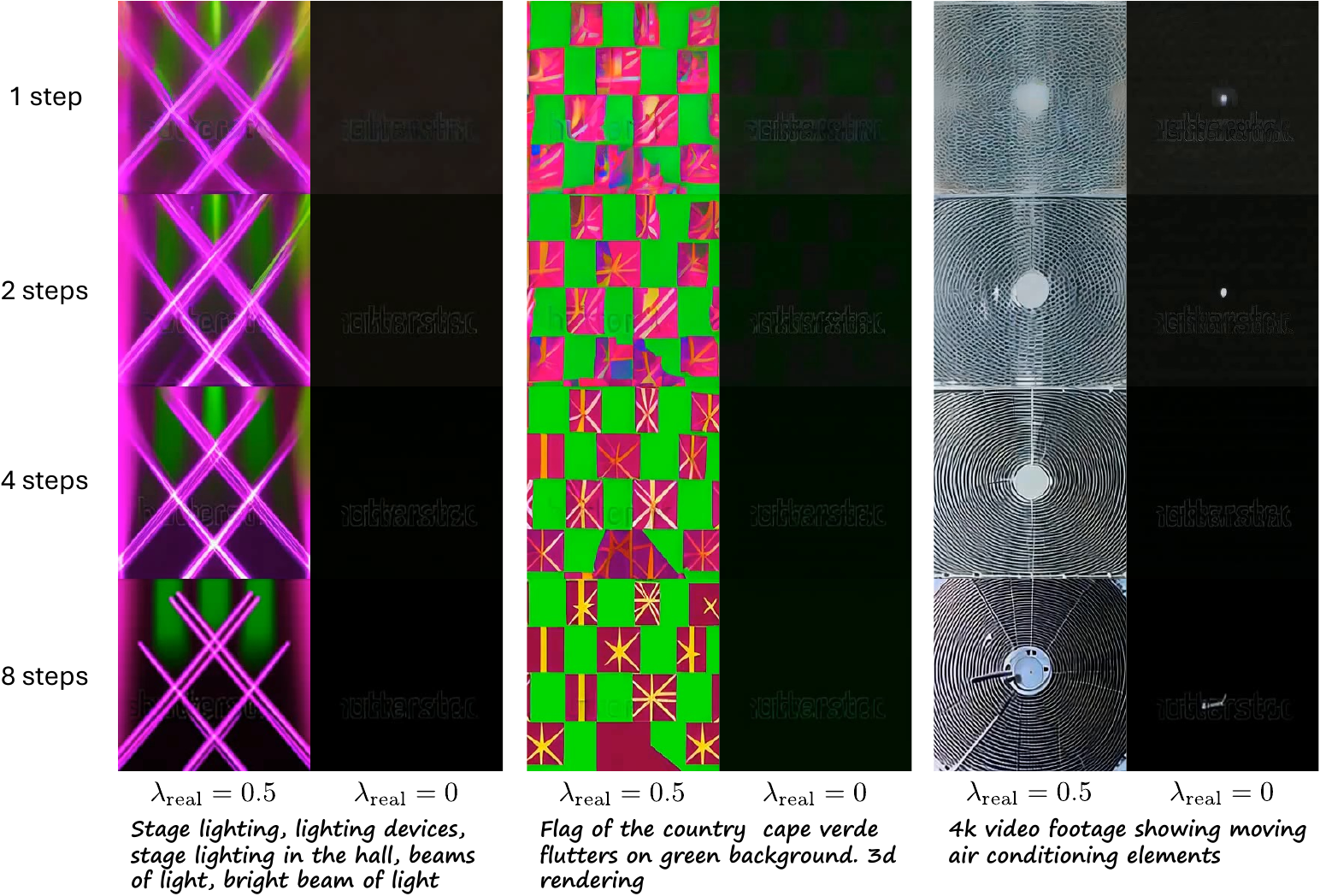}
    \caption{Failure case in mixed trajectory distillation when only generated
    videos are used for ODE trajectories sampling, \ie,
    $\lambda_{\text{real}}=0$.}
    \label{fig:failure-mixed}
\end{figure}

We empirically find that only using generated videos for trajectory sampling
(\ie, $\lambda_{\text{real}}=0$) may sometimes lead to near all-black videos,
accounting for $\sim 5\%$ generated videos.
Some examples are shown in~\cref{fig:failure-mixed}.
Such a result demonstrates the importance of leveraging real video data for the
distillation.

\section{Additional implementation details}
\label{app-sec:additional-details}

\subsection{Image discriminator}

We design our image discriminator based on StyleGAN-T~\cite{sauer2023stylegan}
and ADD~\cite{sauer2023adversarial}.
Specifically, given an input image, the discriminator first extracts features
using a pre-trained, frozen DINOv2~\cite{oquab2023dinov2,darcet2023vitneedreg}
model.
Features are extracted from layers 2, 5, 8, and 11.
These features are then passed through a set of trainable, lightweight
discriminator heads to distinguish images at the patch level.

Following ADD~\cite{sauer2023adversarial}, we use the ViT-S/14 variant for
feature extraction.
To improve video-caption alignment, we also incorporate CLIP embeddings for
additional discriminator conditioning, following
StyleGAN-T~\cite{sauer2023stylegan} and ADD~\cite{sauer2023adversarial}.
For this, we use a CLIP-ViT-g-14 text encoder to compute text embeddings.

\subsection{Training details}

Both teachers use classifier-free guidance~\cite{ho2022classifier}, and we
follow LCM~\cite{luo2023latent} to set a dynamic guidance scale range for distillation.
Specifically, we set the guidance range values to $[5,15]$ for ModelScopeT2V~\cite{wang2023modelscope}, and
$[7,8]$ for AnimateDiff~\cite{guo2023animatediff}.
We follow respective teachers to use $\epsilon$-prediction, and use linear beta scheduling for AnimateDiff following AnimateDiff-Lightning~\cite{lin2024animatediff}.

\end{document}